\documentclass[journal,comsoc]{IEEEtran}
\usepackage[utf8]{inputenc}
\usepackage{kotex} 

\usepackage{multirow}
\usepackage{cite}
\usepackage{dsfont}
\usepackage{algorithmic}
\usepackage{textcomp}
\usepackage{xcolor}
\usepackage{booktabs}
\usepackage{tabularx}

\usepackage[T1]{fontenc}

\ifCLASSINFOpdf
   \usepackage[pdftex]{graphicx}
\else
   \usepackage[dvips]{graphicx}
\fi
\usepackage{amsmath}
\usepackage[cmintegrals]{newtxmath}
\usepackage{bm}

\usepackage{algorithmic}

\usepackage{array}

\ifCLASSOPTIONcompsoc
 \usepackage[caption=false,font=normalsize,labelfont=sf,textfont=sf]{subfig}
\else
 \usepackage[caption=false,font=footnotesize]{subfig}
\fi
\begin{document}
%
\title{FOSNet: An End-to-End Trainable Deep Neural Network for Scene Recognition}
%
%
%

\author{Hongje~Seong,
        Junhyuk~Hyun,
        and~Euntai~Kim,~\IEEEmembership{Members,~IEEE}
\thanks{This research was supported by Next-Generation Information Computing Development Program through the National Research Foundation of Korea(NRF) funded by the Ministry of Science, ICT (NRF-2017M3C4A7069370). \textit{(Corresponding author: Euntai Kim.)}}
\thanks{
H. Seong, J. Hyun, and E. Kim are with the School of Electrical and Electronic Engineering, Yonsei University, Seoul, 120-749, South Korea (e-mail: hjseong@yonsei.ac.kr; jhhyun@yonsei.ac.kr; etkim@yonsei.ac.kr).}
}

%
%

\markboth{}%
{Seong \MakeLowercase{\textit{et al.}}: FOSNet: An End-to-End Trainable Deep Neural Network for Scene Recognition}

%



\maketitle

\begin{abstract}
Scene recognition is an image recognition problem aimed at predicting the category of the place at which the image is taken. In this paper, a new scene recognition method using the convolutional neural network (CNN) is proposed. The proposed method is based on the fusion of the object and the scene information in the given image and the CNN framework is named as FOS (fusion of object and scene) Net. In addition, a new loss named scene coherence loss (SCL) is developed to train the FOSNet and to improve the scene recognition performance. The proposed SCL is based on the unique traits of the scene that the `\textit{sceneness}' spreads and the scene class does not change all over the image. The proposed FOSNet was experimented with three most popular scene recognition datasets, and their state-of-the-art performance is obtained in two sets: 60.14\% on Places 2 and 90.37\% on MIT indoor 67. The second highest performance of 77.28\% is obtained on SUN 397.
\end{abstract}

\begin{IEEEkeywords}
scene recognition, convolutional neural network, fusion network, scene coherence, end-to-end trainable.
\end{IEEEkeywords}

%
\IEEEpeerreviewmaketitle

\section{Introduction}
\label{s1}
%
%
%
%
\IEEEPARstart{S}{cene} recognition is one of the most spotlighted topics in image recognition, applied to image retrieval, autonomous robot, and drone. Many studies have explored the scene recognition; however, most of them have some drawbacks: (1) they consider scene recognition as a simple image recognition problem, and (2) they applied the general CNN or image recognition methods to scene recognition without exploiting the unique traits of scene images \cite{b10,b11,b28,b48}.

In the last few years, some studies have used the scene image traits to improve scene recognition. For example, the scene image traits that a scene image consists of a combination of several objects and the objects in the image possesses much information about the category of the scene was used in \cite{b7,b9}. The traits that the scene categories labels are inherently ambiguous was used in \cite{b5}. The traits of naturalness and openness in scene images were exploited in \cite{b3}. Unfortunately, however, most of the existing methods used scene traits to extract scene features manually, and only a limited number of studies trained the features of scene images with the scene traits and extracted them automatically.

In this paper, a new scene recognition framework named FOSNet is proposed. The FOSNet improves the scene recognition performance by exploiting the unique traits of the scene images; the traits that FOSNet uses are summarized as follows:

\begin{enumerate}
\item[(T1)] The \textit{sceneness} spreads all over the image, and the classes of the scene are the same to the entire image. This is contrary in the object images, where the \textit{objectness} appears only at specific locations in an image. Thus, the classes of the objects change from patch to patch in the same image. This trait is named scene coherence (SC) in the scene image.

\item[(T2)] Various objects can appear on the scene image, and when a specific object is found, the scene image belongs to a particular class. For example, if a chair is found, the corresponding image is unlikely to be a `mountain' or a `playground', but it is likely to be a `meeting room' or a `classroom'.
\end{enumerate}

Based on (T1), a scene coherence loss (SCL) is developed to train the FOSNet. This SCL favors the case in which the class of the scene is the same all over the image and it penalizes the case in which the classes of the adjacent patches in a single image are different from each other. Based on (T2), the correlative context gating (CCG) is developed to combine the object and scene features in a given image. The fusion of the object and scene features in the image is not new but it has been reported in the previous works \cite{b7,b9}. However, the proposed CCG is completely different from the previous works, as the previous works combined the object and scene features in the image by simply concatenating two features and increasing the feature dimension. This kind of simple fusion or concatenation could increase the redundancy among features and might degrade the recognition performance. On the other hand, the CCG combines the two features in an efficient manner, applying the attention concept to the fusion of two features. The CCG is an improved version of the class conversion matrix (CCM) \cite{b21}, and it also determines the relative importance of the features during training. 

The contributions of FOSNet are as follows:

\begin{enumerate}
\item	The traits of scene coherence (SC) in a scene image are defined, and a new loss SCL is developed based on the trait. The SCL is the \textit{only} loss specialized for scene recognition and it is applied to train the FOSNet. 

\item	A new fusion framework named CCG is proposed to combine the object and scene features from the image. Unlike the previous fusion methods in which the two features are simply concatenated and the classifier is designed for the features, the CCG selects important features and fuses the two sets of features effectively for training.
\end{enumerate}

The rest of the paper is organized as follows: Section \ref{s2} provides a brief review of the related studies. Section \ref{s3} explained FOSNet in detail. Section \ref{s4} applies the FOSNet to three benchmark problems, and the performance of the FOSNet is demonstrated through experimentation. Section \ref{s5} conducts some ablation study to verify the value of our proposed SCL and CCG. Section \ref{s6} concludes the paper.

\section{Related Works}
\label{s2}

In this section, we review previous works on scene recognition with an emphasis on (1) the application of scene traits and (2) a combination of the object and scene information.

\subsection{Traits in Scene Image}
\label{s21}

Using object information in an image is the most utilized scene traits for scene recognition \cite{b7,b9,b21}. When a particular object appears in an image, the chance of the image belonging to a certain category associated with the object increases. For example, if a TV is detected, the chance of the image being in a living room increases. In the previous works, the objects features were used for scene recognition instead of detecting the objects directly. To extract the object features, large image datasets for object recognition are used and they are ImageNet \cite{b19}, PASCAL visual object classes (VOC) \cite{b42}, or Microsoft COCO \cite{b43}.

Previous studies also used other scene traits: The analysis of object scales in the scene images was utilized in \cite{b7} and \cite{b6}. In \cite{b9,b13,b14,b15}, the number of CNN input patches was adjusted by considering several objects in the scene image. To capture recurring visual elements and salient objects in scene recognition, the deformable part-based model (DPM) was utilized in \cite{b2}. In addition, the traits that features appearing in each image region within scene images are all similar was used in \cite{b1}. A super category was proposed to solve the problem that the scene categories have label ambiguity in \cite{b5}. A deep gaze shifting kernel was developed to distinguish sceneries from different categories in \cite{b12}. As such, the traits of the scene image are very diverse, and there seem to be still many available unused scene traits in scene recognition studies.

\subsection{Feature Fusion in Scene Recognition}
\label{s22}

In order to combine scene and object features for scene recognition, the effective feature fusion is great importance and several fusion methods have been reported. For example, the two features extracted by two different CNNs were combined at feature level by summing or concatenating the features in \cite{b3,b4,b5,b6,b7,b8,b9,b12,b13,b14,b15,b16,b17}. Then, the classical classifiers such as support vector machine (SVM) \cite{b49} was applied to the fused features. Unfortunately, these methods have some drawbacks that they cannot be trained in an end-to-end manner. Moreover, the simple summation or concatenation might degrade the recognition performance owing to redundancy in two feature sets.

Recently, the CCM has learned a relationship between an object feature and scene feature through training and converted object features into scene features \cite{b21}. By doing so, the two features in different domains are transformed into the same domain, and the fusion is performed through the element-wise sum operation. In the CCM, objects in the images do not need to be labeled in the scene recognition dataset; only a pre-trained CNN trained on the object recognition dataset is enough and the relationship between scene and objects is trained in a weakly supervised way. In this paper, a new fusion method named correlative context gating (CCG) is proposed. The CCG is an extended version of the CCM and it generates more accurate scene features than CCM using the concept of attention.

\section{Proposed Methods}
\label{s3}

\begin{figure}[!t]
\centering
\includegraphics[width=0.95\linewidth]{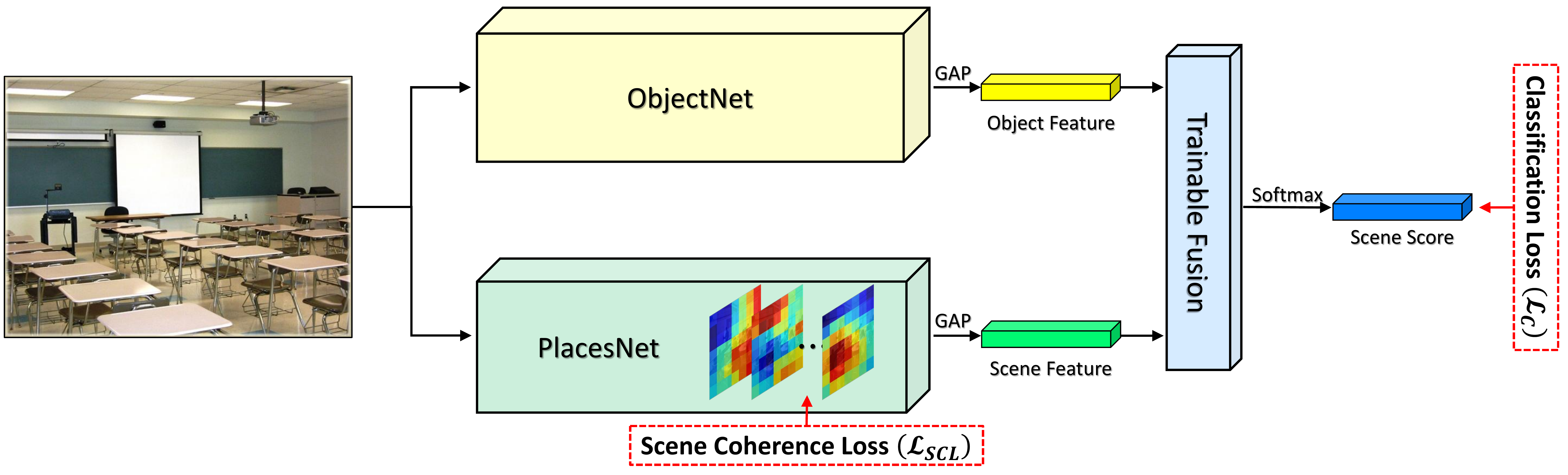}
\caption{An overall architecture of FOSNet.}
\label{fig1}
\end{figure}

In this section, a new scene recognition network named FOSNet is proposed. The overall FOSNet structure is shown in Fig. \ref{fig1}. As shown in the figure, FOSNet has two input streams. In the upper stream named ObjectNet, the features of the objects in the scene images are extracted. In the lower stream named PlacesNet, scene features are extracted. In a trainable fusion module, two streams of features are fused into a combined feature for scene recognition. The FOSNet consists of ObjectNet, PlacesNet, and trainable fusion modules, and all networks can be trained in an end-to-end manner. The three subnets are explained in detail in the subsequent subsections.

\subsection{ObjectNet}
\label{s31}
\begin{figure}[!t]
\centering
\includegraphics[width=0.95\linewidth]{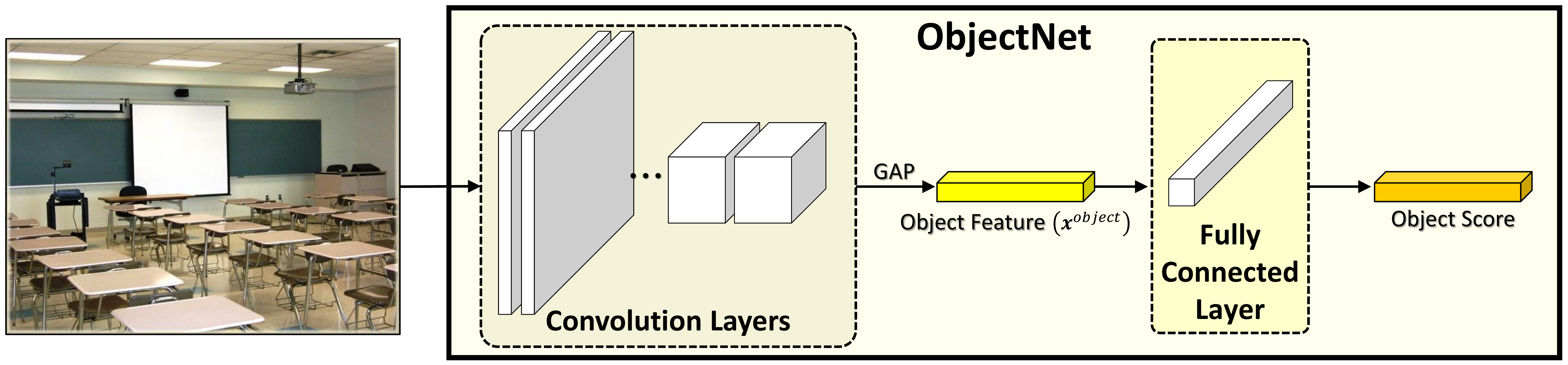}
\caption{Structure of ObjectNet.}
\label{fig2}
\end{figure}

Based on the scene traits (T2), information about the objects that appear in the scene is exploited in FOSNet. To obtain a highly discriminating object descriptor, ObjectNet is utilized in the upper stream of Fig. \ref{fig1} to extract a feature of the objects in a scene image. As the ObjectNet, the popular CNN models \cite{b23,b26,b27,b28} pre-trained on ImageNet \cite{b19} are used, as shown in Fig. \ref{fig2}. An object feature extracted through ObjectNet is fed into the trainable fusion module. In the structure given in Fig. \ref{fig2}, not only the object feature but also the object score can be fed into the trainable fusion module. Detailed description of the fusion level is given in Section \ref{s33}.

\subsection{PlacesNet}
\label{s32}

\begin{figure}[!t]
\centering
\subfloat[]{
    \includegraphics[width=0.95\linewidth]{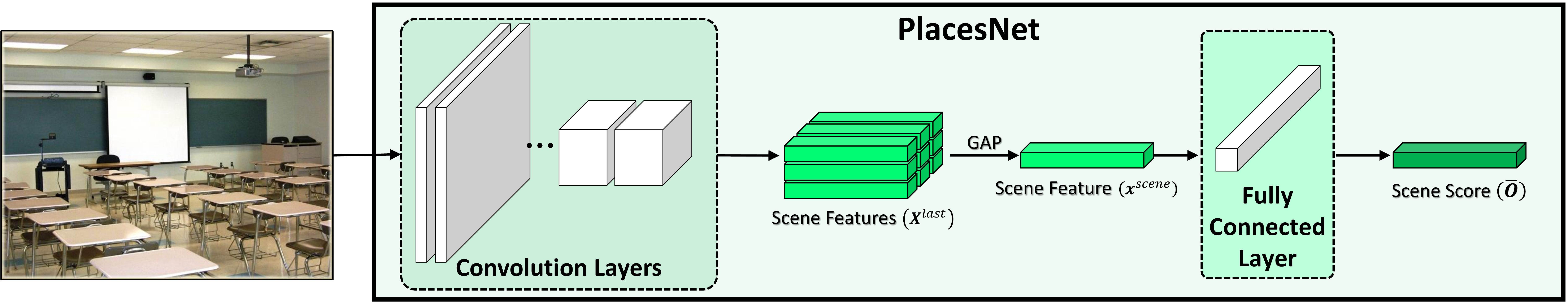}
    \label{fig3a}
}

\subfloat[]{
    \includegraphics[width=0.95\linewidth]{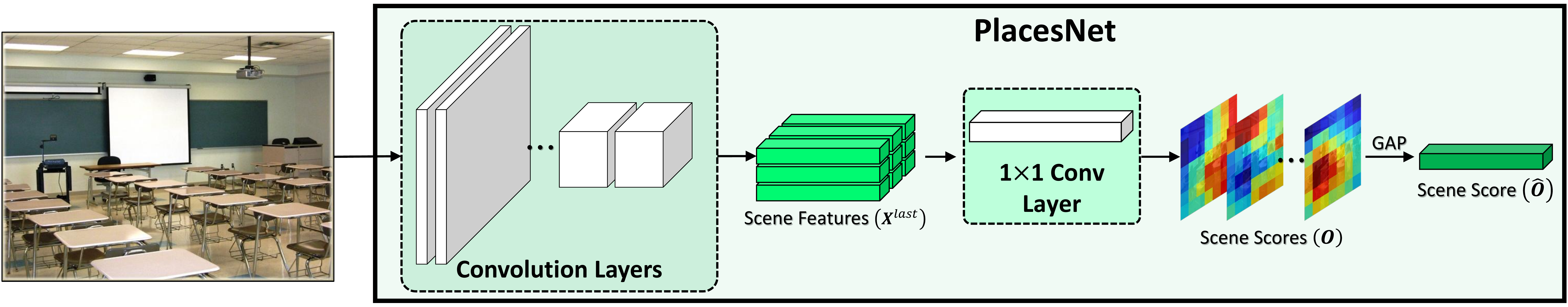}
    \label{fig3b}
}
\caption{Structures of PlacesNet. (a) Vanilla CNN structure; PlacesNet should be applied multiple times to compute the SCL. (b) A new structure in which SCL can be computed by applying PlacesNet only once.}
\label{fig3}
\end{figure}

\begin{figure}[!t]
\centering
\includegraphics[width=0.95\linewidth]{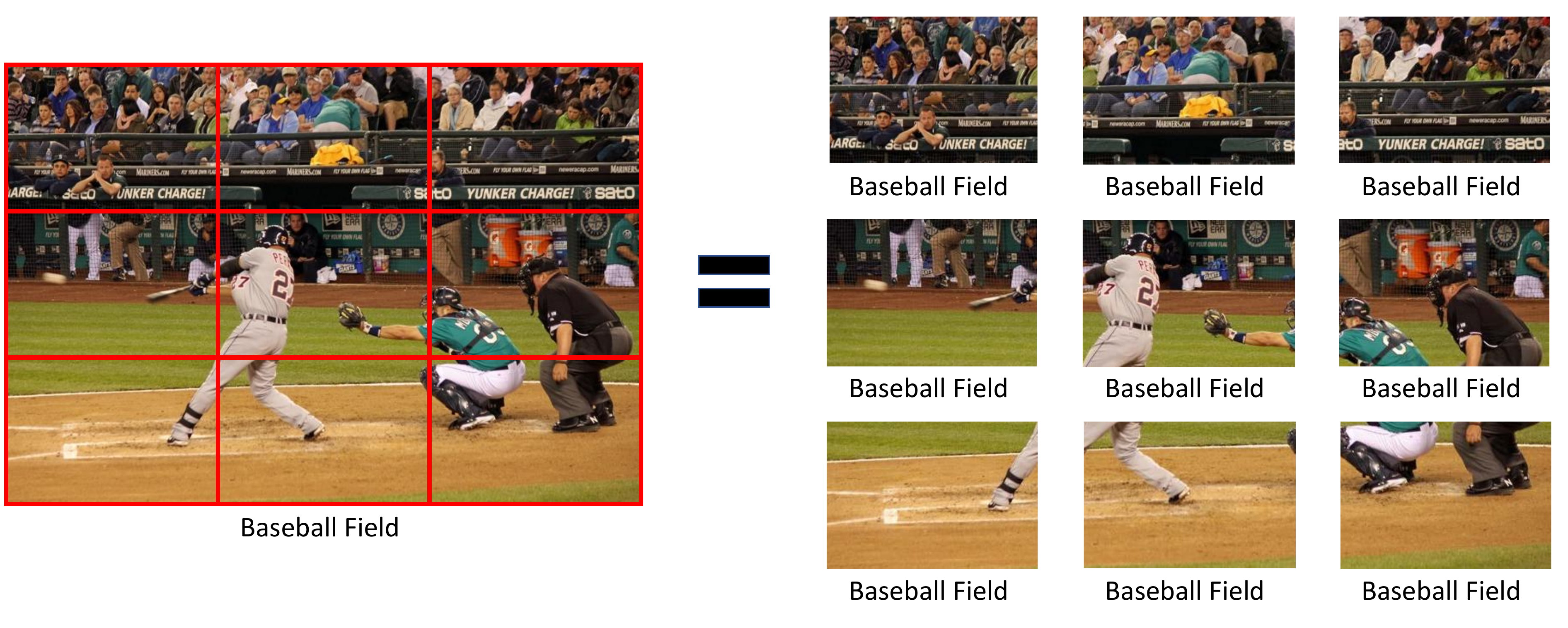}
\caption{Scene coherence in a scene image. Even if a scene image is divided into multiple grids, each grid cell represents the same scene.}
\label{fig4}
\end{figure}

PlacesNet is another CNN model and it extracts a scene feature from an image. The PlacesNet is pre-trained using Places 2 \cite{b20} and its structure is the same as that of the ObjectNet, as shown in Fig. \ref{fig3}(a). To train the PlacesNet, scene coherence loss (SCL) is developed in this paper. The SCL is a new loss tailored for scene recognition, and it is based on the scene trait that \textit{objectness focused on specific parts of an image, whereas sceneness unfocused on specific parts. However, it spreads all over the image. In particular, the class of the scene is unchanged over the image.} This trait is named the coherence in the scene of an image, and the SCL embodies this trait into a single loss. For example, let us consider an image shown in Fig. \ref{fig4}. When the whole image is divided into nine grids, all nine grids cannot have different scene classes and all of them have the same scene class of a baseball field. That is, the scene class should be \textit{coherent} all over the image.

\begin{figure}[!t]
\centering
\includegraphics[width=0.95\linewidth]{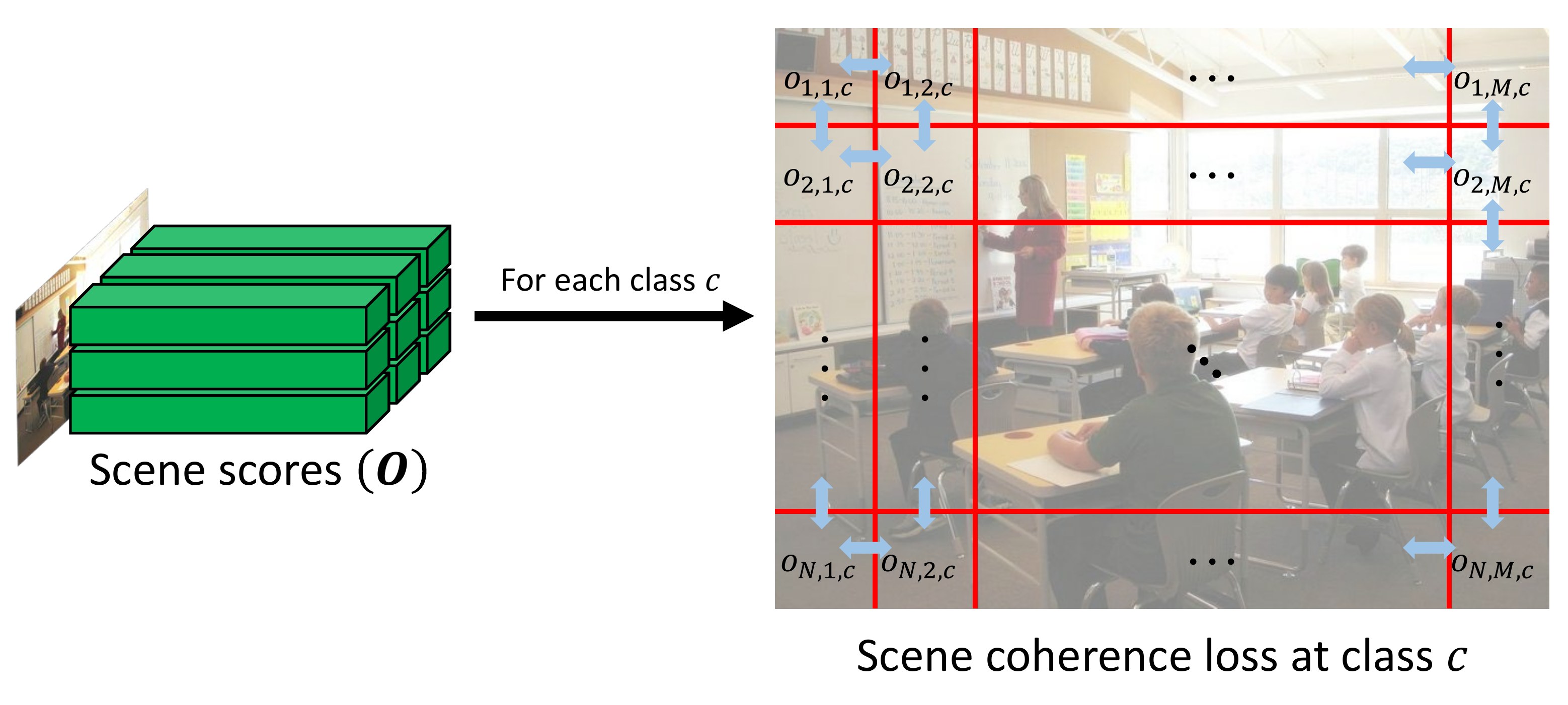}
\caption{Visualization of scene coherence loss (SCL).}
\label{fig5}
\end{figure}

The scene coherence is a unique trait of the scene image and it is formulated into a new loss SCL:
\begin{equation}
\begin{aligned}
{{\cal L}_{SCL}} = {} & \frac{1}{C}\sum\limits_{c = 1}^C {\frac{1}{{\left( {N - 1} \right)M + N\left( {M - 1} \right)}}} \\
& \qquad \times \left( {\sum\limits_{n = 1}^{N - 1} {\sum\limits_{m = 1}^M {{{\left( {{o_{n + 1,m,c}} - {o_{n,m,c}}} \right)}^2}} }} \right. \\
& \qquad \qquad \left. + {\sum\limits_{n = 1}^N {\sum\limits_{m = 1}^{M - 1} {{{\left( {{o_{n,m + 1,c}} - {o_{n,m,c}}} \right)}^2}} } } \right)
\label{eq1}
\end{aligned}
\end{equation}
where $N$ and $M$ are the numbers of grid cells in the vertical and horizontal directions, respectively; $C$ is the number of classes; ${o_{n,m,c}}$ denotes the classification result for the class $c$ in the grid cell $\left( n,m \right)$, as shown in Fig. \ref{fig5}. As stated, the SCL defined in Eq. \ref{eq1} favors the case in which all the grids have the same scene class, whereas it penalizes the case in which the adjacent grids have the different scene classes.

Here, when we apply SCL in Eq. \ref{eq1} to the PlacesNet training, a difficulty arises; The PlacesNet should be applied to $N \times M$ grid cells separately and repeatedly and it leads to the waste of computation time. To resolve this inefficiency, the PlacesNet in Fig. \ref{fig3}(a) is converted into the form of a fully convolutional network, as shown in Fig. \ref{fig3}(b). The conversion is motivated by class activation map (CAM) \cite{b25} and it can be applied to any CNNs, in which the last layers are global average pooling (GAP) followed by fully connected (FC) layers. In the PlacesNet, the input image with the size of $224 \times 224$ is reduced to $7 \times 7$ feature map after going through five pooling operations in convolutional layers. Then, the scene scores for each $7 \times 7$ grid cell is obtained by replacing the last GAP-FC sequence with $1 \times 1$ convolution. Then, a scene score for the entire image is computed by applying the GAP to the tensors obtained from $1 \times 1$ convolution.

Interestingly, it can be shown that the PlacesNet with the GAP followed by FC shown in Fig. \ref{fig3}(a) outputs the same result with the converted version with the $1 \times 1$ convolution followed by GAP shown in Fig. \ref{fig3}(b). With slightly relaxed notation, the feature tensor extracted from the PlacesNet in Fig. \ref{fig3}(a) is represented into
\begin{equation}
\begin{aligned}
  {{\bm{X}}^{last}} = {} & \left( {x_{n,m,d}^{last}} \right) \\ 
   = & \left( {\begin{array}{*{20}{c}}
  {{\bm{x}}_{1,1,1:D}^{last}}&{{\bm{x}}_{1,2,1:D}^{last}}& \cdots &{{\bm{x}}_{1,M,1:D}^{last}} \\ 
  {{\bm{x}}_{2,1,1:D}^{last}}&{{\bm{x}}_{2,2,1:D}^{last}}& \cdots &{{\bm{x}}_{2,M,1:D}^{last}} \\ 
   \vdots & \vdots &{}& \vdots  \\ 
  {{\bm{x}}_{N,1,1:D}^{last}}&{{\bm{x}}_{N,1,1:D}^{last}}& \cdots &{{\bm{x}}_{N,M,1:D}^{last}} 
\end{array}} \right) \\ 
  \, & \in {\mathbb{R}^{N \times M \times D}} \\ 
\end{aligned} 
\label{eq2}
\end{equation}
where $N \times M$ is the feature map size extracted from convolution layers in Fig. \ref{fig3}(a); $D$ is the number of output channels of last convolution layer; ${\bm{x}}_{n,m,1:D}^{last} \in {\bm{R}^D}$ denotes a feature vector at position $\left( n,m \right)$ of ${\bm{X}}^{last}$; $1:D$ in the third axis of ${\bm{x}}_{n,m,1:D}^{last}$,   is a collection of all the elements accumulated over $D$ channels and it is actually a vector. Let the trainable parameters ${\bm{W}} = \left( {{w_{c,d}}} \right) \in {\mathbb{R}^{C \times D}}$ and ${\bm{b}} = \left( {{b_c}} \right) \in {\mathbb{R}^{C}}$ be weight and bias, respectively, for the FC layer of the model in Fig. \ref{fig3}(a), and $\overline {\bm{O}}  \in {\mathbb{R}^C}$, $\widehat {\bm{O}} \in {\mathbb{R}^C}$ be the classification results of the models in Figs. \ref{fig3}(a) and (b), respectively. Then, we can prove that $\overline {\bm{O}}$ and $\widehat {\bm{O}}$ are the same by
\begin{equation}
\begin{aligned}
  \overline {\bm{O}}  = {} & \operatorname{FC} \left( {\operatorname{GAP} \left( {{{\bm{X}}^{last}}} \right),{\bm{W}},{\bm{b}}} \right) \\ 
   = & \operatorname{FC} \left( {\frac{1}{{NM}}\sum\limits_{n = 1}^N {\sum\limits_{m = 1}^M {{\bm{x}}_{n,m,1:D}^{last}} } ,{\bm{W}},{\bm{b}}} \right)\quad  \\ 
   = & {\bm{W}}\left( {\frac{1}{{NM}}\sum\limits_{n = 1}^N {\sum\limits_{m = 1}^M {{\bm{x}}_{n,m,1:D}^{last}} } } \right) + {\bm{b}}\quad  \\ 
   = & \left( {\frac{1}{{NM}}\sum\limits_{n = 1}^N {\sum\limits_{m = 1}^M {\left( {{\bm{Wx}}_{n,m,1:D}^{last} + {\bm{b}}} \right)} } } \right)\quad  \\ 
   = & \left( {\frac{1}{{NM}}\sum\limits_{n = 1}^N {\sum\limits_{m = 1}^M {{{\operatorname{Conv} }^{1 \times 1}}{{\left( {{{\bm{X}}^{last}},{\bm{W}},{\bm{b}}} \right)}_{n,m}}} } } \right) \\ 
   = & \operatorname{GAP} \left( {{{\operatorname{Conv} }^{1 \times 1}}\left( {{{\bm{X}}^{last}},{\bm{W}},{\bm{b}}} \right)} \right) \\ 
   = & \widehat {\bm{O}} \\ 
\label{eq3}
\end{aligned}
\end{equation}
where
\begin{equation}
\begin{aligned}
  {} & {\operatorname{Conv} ^{1 \times 1}}\left( {{{\bm{X}}^{last}},{\bm{W}},{\bm{b}}} \right) \hfill \\
   & \qquad = \left( {\begin{array}{*{20}{c}}
  {{\bm{Wx}}_{1,1,1:D}^{last} + {\bm{b}}}& \cdots &{{\bm{Wx}}_{1,M,1:D}^{last} + {\bm{b}}} \\ 
   \vdots &{}& \vdots  \\ 
  {{\bm{Wx}}_{N,1,1:D}^{last} + {\bm{b}}}& \cdots &{{\bm{Wx}}_{N,M,1:D}^{last} + {\bm{b}}} 
\end{array}} \right) \\
& \qquad \quad \in {\mathbb{R}^{N \times M \times C}} \hfill \\ 
\end{aligned} 
\end{equation}
is a tensor obtained by applying $1 \times 1$ convolution with weights $\left( {{\bm{W}},{\bm{b}}} \right)$ to input ${{\bm{X}}^{last}}$, and it is also the classification results for each grid cell shown in Fig. \ref{fig3}(b). Since $\overline {\bm{O}}$ and $\widehat {\bm{O}}$ have the same values, the model in Fig. \ref{fig3}(b) performs the same classification as the one in Fig. \ref{fig3}(a) and it has an advantage of obtaining classification results ${\bm{O}} \triangleq {\operatorname{Conv} ^{1 \times 1}}\left( {{{\bm{X}}^{last}},{\bm{W}},{\bm{b}}} \right) \in {\mathbb{R}^{N \times M \times C}}$ for all grid cells without applying PlacesNet to all grid cells repeatedly. Here, classification error is defined using the cross-entropy loss and it is denoted by 
\begin{equation}
{\mathcal{L}_C} =  - \sum\limits_c {{y_c}\log \left( {\frac{{\exp \left( {{{\widehat o}_c}} \right)}}{{\sum\limits_c {\exp \left( {{{\widehat o}_c}} \right)} }}} \right)}  ,
\label{eq5}
\end{equation}
where
\begin{equation}
\begin{aligned}
\widehat {\bm{O}} = {} & \left( {{{\widehat o}_c}} \right) \hfill \\
   = & GAP\left( {{o_{n,m,c}}} \right) \hfill \\
   = & \left( {\frac{1}{{NM}}\sum\limits_{n = 1}^N {\sum\limits_{m = 1}^M {{o_{n,m,c}}} } } \right) \in {\mathbb{R}^C} \hfill \\ 
\label{eq6}
\end{aligned}
\end{equation}
is a vector of classification results for $C$ classes and it is obtained by applying GAP to the result of Eq. (4); ${\bm{Y}} = \left( {{y_c}} \right) \in {\left\{ {0,1} \right\}^C}$ denotes the ground truth of the class of the given scene image and it is represented by a one-hot vector.

Then, the total training loss ${\mathcal{L}_{total}}$ is defined as a summation of the proposed SCL ${\mathcal{L}_{SCL}}$ and classification loss ${\mathcal{L}_{C}}$, and it is represented by
\begin{equation}
{\mathcal{L}_{total}} = {\mathcal{L}_C} + \gamma {\mathcal{L}_{SCL}}
\label{eq7}
\end{equation}
where $\gamma$ denotes the SCL rate and controls the relative weight between SCL and the classification loss.

\begin{figure}[!t]
\centering
\includegraphics[width=0.95\linewidth]{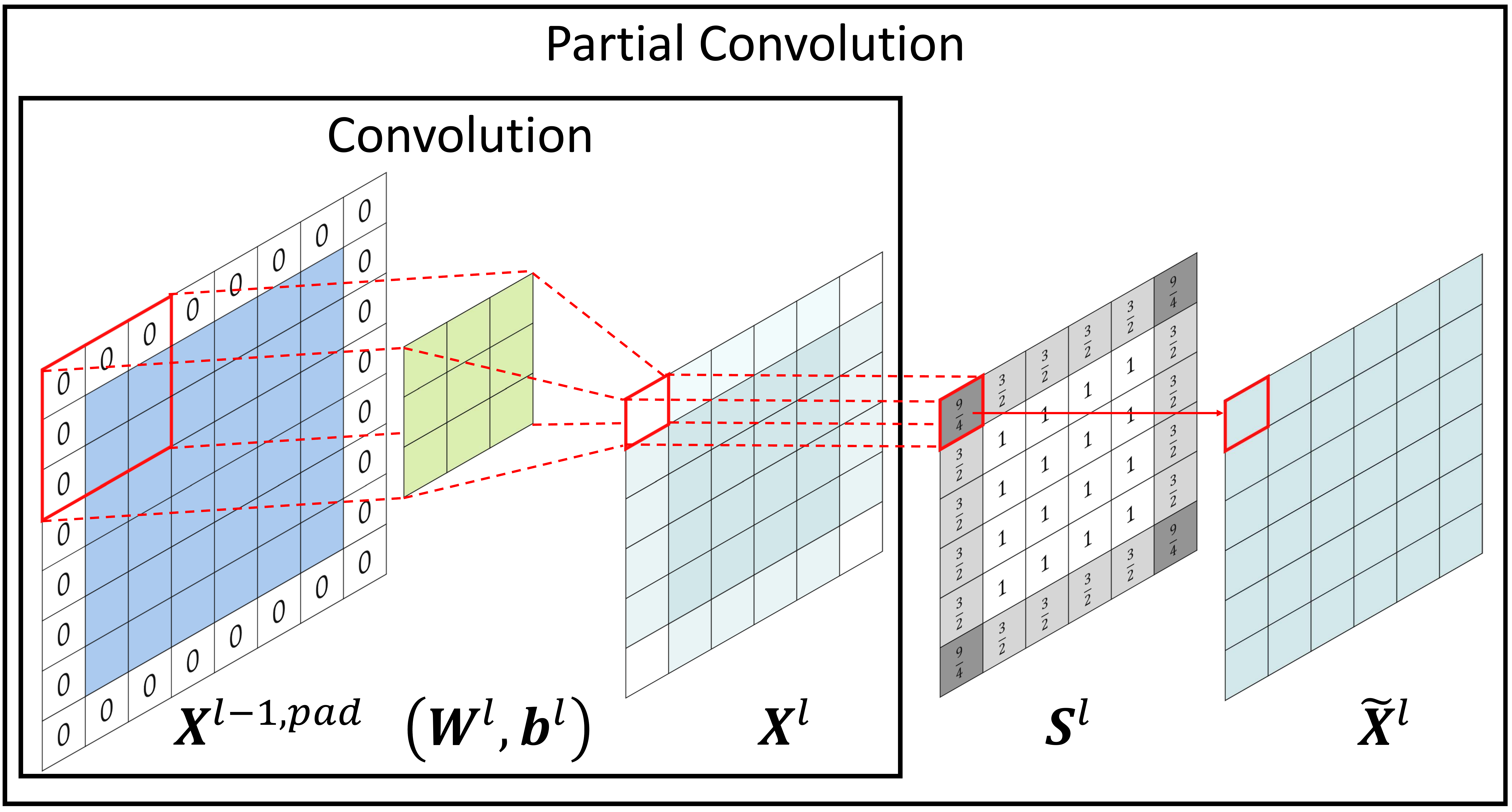}
\caption{Illustration of convolution with zero padding and partial convolution.}
\label{fig6}
\end{figure}

Another key feature of PlacesNet is that partial convolution \cite{b24} is applied to all convolution layers. In vanilla convolution with zero padding, boundary of the image is filled with zeros and the vanilla convolution is applied, as shown in Fig. \ref{fig6}. Using the padded input ${\bm{x}}_{n,m,1:{D^l}}^{l - 1,pad} = \left( {x_{n,m,{d^l}}^{l - 1,pad}} \right) \in {\mathbb{R}^{{D^l}}}$, the output vector ${\bm{x}}_{n,m,1:{D^{l + 1}}}^l$ of the $l$-th layer of vanilla convolutions at the position $\left( {n,m} \right)$ is computed as follows: 
\begin{equation}
x_{n,m,{d^l}}^l = \sum\limits_{i = 1}^{{H^l}} {\sum\limits_{j = 1}^{{W^l}} {W_{i,j,{d^{l - 1}},{d^l}}^lx_{n + i,m + j,{d^l}}^{l - 1,pad}} }  + b_{{d^l}}^l
\label{eq8}
\end{equation}
where ${{\bm{W}}^l} = \left( {W_{i,j,{d^{l - 1}},{d^l}}^l} \right) \in {\mathbb{R}^{{H^l}\times{W^l}\times{D^{l - 1}}\times{D^l}}}$ and ${{\bm{b}}^l} = \left( {b_{{d^l}}^l} \right) \in {\mathbb{R}^{{D^l}}}$ are the filter weights of the $l$-th layer; ${H^l}$ and ${W^l}$ are height and width of filter size respectively; ${D^l}$ is the number of output channels of the $l$-th layer. Here, it can be seen that vanilla convolution ${{\bm{X}}^l} = \left( {{\bm{x}}_{n,m,1:{D^l}}^l} \right)$ around the boundary of the feature might not be as accurate as that inside the feature since the vanilla convolution should include many zero paddings. Recently, a lot of convolution layers are connected sequentially, and then the performance deterioration of the boundary of the given image becomes worse. For other general classification CNNs, the accuracy degradation around the boundary of the image might not be important because GAP is used before the classification. In FOSNet, however, the SCL is used as a loss and the classification accuracy around the boundary of the image is as important as that inside the image. Thus, the partial convolution proposed in \cite{b24} is used in FOSNet.

The structure of partial convolution \cite{b24} is given in Fig. \ref{fig6}. The scaling mask ${{\bm{S}}^l} = \left( {S_{n,m}^l} \right)$ is multiplied with the convolution ${{\bm{X}}^l}$, and the output of the partial convolution is computed by 
\begin{equation}
\widetilde x_{n,m,{d^l}}^l = S_{n,m}^l\sum\limits_{i = 1}^{{H^l}} {\sum\limits_{j = 1}^{{W^l}} {W_{i,j,{d^{l - 1}},{d^l}}^lx_{n + i,m + j,{d^l}}^{l - 1,pad}} }  + b_{{d^l}}^l
\label{eq9}
,
\end{equation}
where
\begin{equation}
S_{n,m}^l = \frac{{{H^l}{W^l}}}{{{H^l}{W^l} - \sum\limits_{i = 1}^{{H^l}} {\sum\limits_{j = 1}^{{W^l}} {{\mathds{1}}_{n + i,m + j}^{l;pad}} } }}
\label{eq10}
,
\end{equation}

where $S_{n,m}^l$ is scaling factor of output feature $\widetilde x_{n,m,{d^l}}^l$; ${\mathds{1}}_{n,m}^{l;pad}$ is 1 if position $\left( {n,m} \right)$ of the input feature $x_{n,m,d}^{l - 1,pad}$ is zero padded position, otherwise 0. Therefore, partial convolution adjusts for the varying amount of valid inputs by scaling, and it likely increases the accuracy of the near image boundary.

The partial convolution is a good match with the SCL and it will be shown that the combination enhances the classification accuracy significantly. The analysis will be given in Section \ref{s5}.

\subsection{Fusion of Object Feature and Scene Feature}
\label{s33}

\begin{figure}[!t]
\centering
\subfloat[]{
    \includegraphics[height=0.11\textheight]{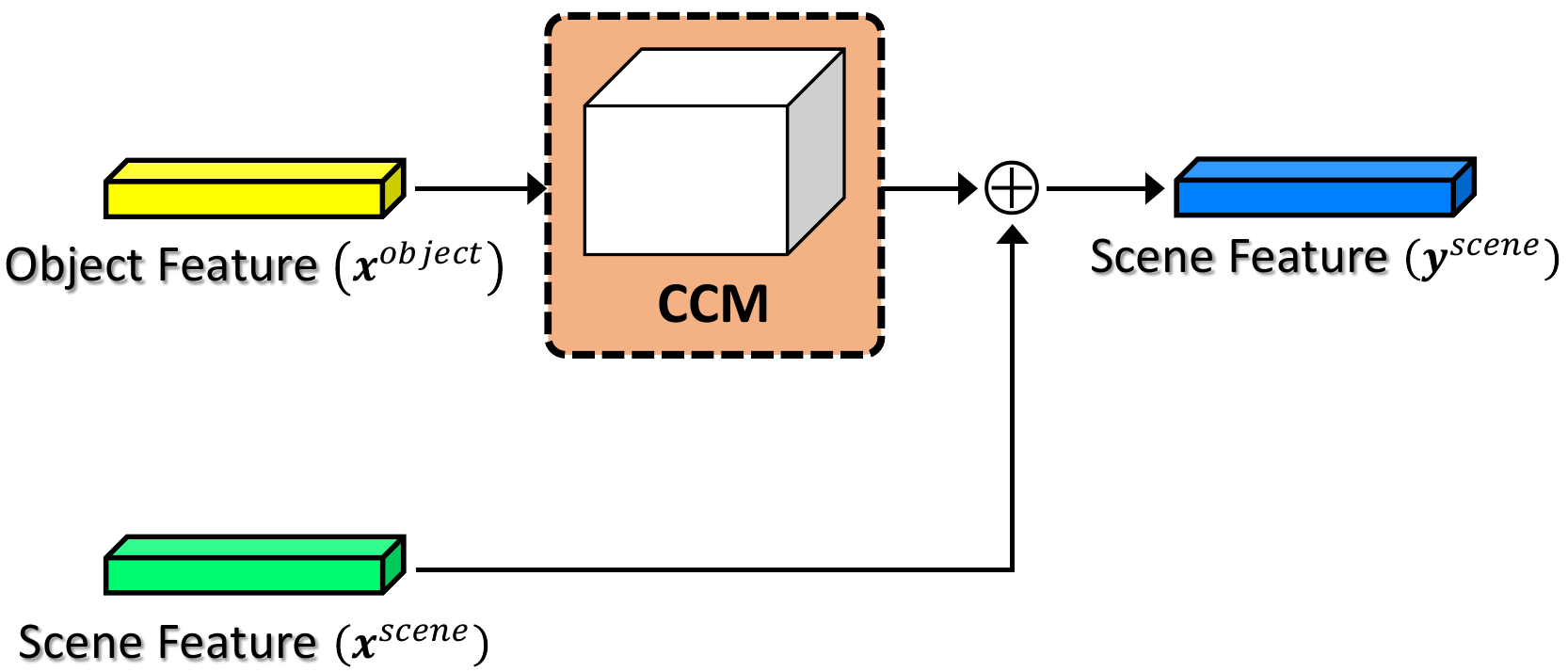}
    \label{fig7a}
}

\subfloat[]{
    \includegraphics[height=0.11\textheight]{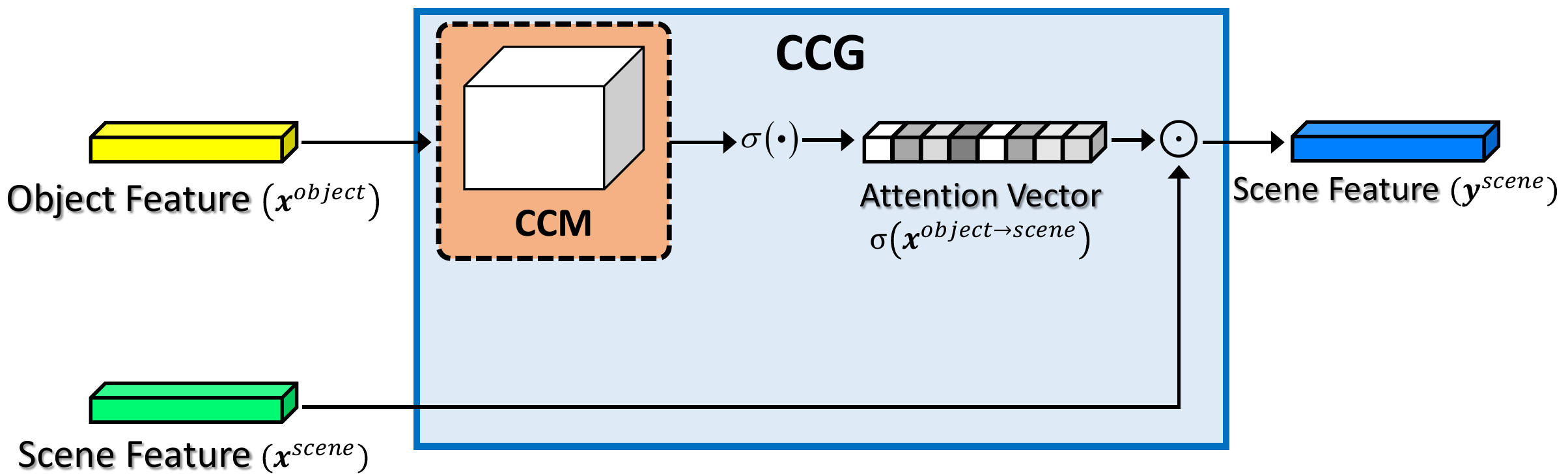}
    \label{fig7b}
}

\subfloat[]{
    \includegraphics[height=0.11\textheight]{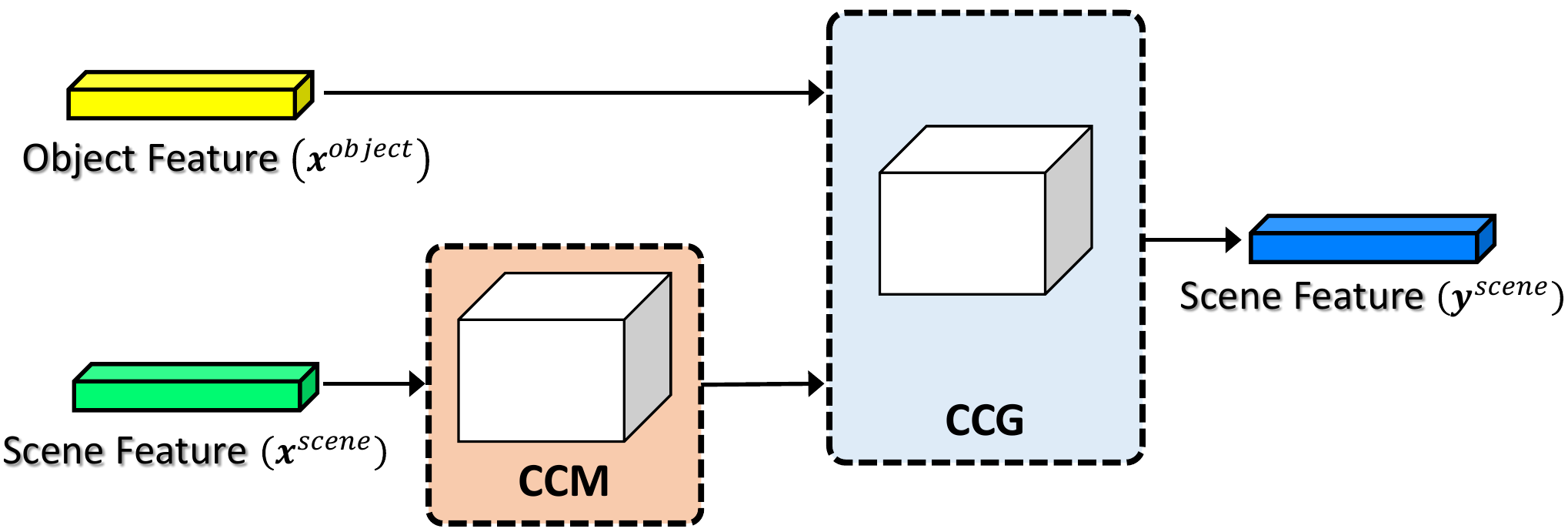}
    \label{fig7c}
}
\caption{Trainable fusion modules with object feature and scene feature. (a) CCM \cite{b21}; (b) CCG; (c) mixed CCM-CCG.}
\label{fig7}
\end{figure}

In this subsection, a new fusion module CCG is proposed. The CCG combines object feature $\bm{x}^{object}$ containing information of objects in the image with scene feature $\bm{x}^{scene}$. $\bm{x}^{object}$ is extracted from ObjectNet in Fig. \ref{fig2}, while $\bm{x}^{scene}$ is extracted from PlacesNet that is trained using SCL in Fig. \ref{fig3}(b). The CCG is based on a scene traits that \textit{when a specific object in an image is found, the scene is very likely to belong to a particular class associated with the object}. The CCG is inspired by context gating \cite{b22} and the CCM \cite{b21}. The concept of CCG is depicted in Fig. \ref{fig7}.


Using CCM \cite{b21}, CCG converts an object feature into a scene feature and outputs a pseudo scene feature ${{\bm{x}}^{object \to scene}}$. Then, an attention map is generated by applying a sigmoid function to ${{\bm{x}}^{object \to scene}}$. The scene feature ${{\bm{x}}^{scene}}$ from PlacesNet is multiplied by the generated attention map $\sigma \left( {{{\bm{x}}^{object \to scene}}} \right)$ in element-wise manner, and a new scene feature ${{\bm{y}}^{scene}}$ is obtained by
\begin{equation}
\begin{aligned}
  {{\bm{y}}^{scene}} = {} & \sigma \left( {{{\bm{x}}^{object \to scene}}} \right) \odot {{\bm{x}}^{scene}} \\ 
   = & \sigma \left( {{\bm{W}}{{\bm{x}}^{object}} + {\bm{b}}} \right) \odot {{\bm{x}}^{scene}} \\ 
\end{aligned} 
\label{eq11}
\end{equation}
where $\odot$ denotes element-wise multiplication; $\bm{W}$ and $\bm{b}$ are the trainable parameters; ${{\bm{x}}^{object \to scene}}$ is a pseudo scene feature obtained by converting the object feature into the scene feature through CCM, and $\sigma \left( x \right) = \frac{1}{{1 + \exp \left( { - x} \right)}}$ is a sigmoid function.
The structure of CCG is motivated by context gating \cite{b22}. The context gating transforms the input feature into a new feature using a self-gating mechanism, and it demonstrated significant improvements in video understanding tasks. Motivated by context gating, CCG selectively activates the channels of scene feature ${{\bm{x}}^{scene}}$. The selective activation is carried out by applying a gating mechanism at the object feature ${{\bm{x}}^{object}}$, which are relevant to scene recognition. Here CCM \cite{b21} is applied to the object feature ${{\bm{x}}^{object}}$, and it converts ${{\bm{x}}^{object}}$ into a pseudo scene feature ${{\bm{x}}^{object \to scene}}$ to modify the context gating concept of the self-gating mechanism into the correlative-gating mechanism. The structure of CCG is shown in Fig. \ref{fig7}(b). As applied in the batch normalization (BN) \cite{b32} at the CCM in \cite{b21}, batch normalization can be applied to CCG as in 
\begin{equation}
{{\bm{y}}^{scene}} = \sigma \left( {\operatorname{BN} \left( {{\bm{W}}{{\bm{x}}^{object}}} \right)} \right) \odot {{\bm{x}}^{scene}}
\label{eq12}
.
\end{equation}

Another variation, a mixed CCM-CCG, can also be considered. Since PlacesNet is pre-trained using Places 2 dataset \cite{b20}, performance degradation might occur when PlacesNet is applied to scene recognition datasets other than Places 2 (e.g., SUN397 \cite{b33}, MIT 67\cite{b34}). To obtain ${{\bm{x}}^{scene \to scen{e^{target}}}}$, CCM converts the scene feature extracted from the PlacesNet to the feature suitable for the target scene dataset. Then, the converted ${{\bm{x}}^{scene \to scen{e^{target}}}}$ and object features are fused using CCG. In this case, the mixed CCM-CCG proceeds as follows:
\begin{equation}
\begin{aligned} 
  {{\bm{y}}^{scene}} = {} & \sigma \left( {{{\bm{x}}^{object \to scene}}} \right) \odot {{\bm{x}}^{scene \to scen{e^{target}}}} \\ 
   = & \sigma \left( {{{\bm{W}}^1}{{\bm{x}}^{object}} + {{\bm{b}}^1}} \right) \odot \left( {{{\bm{W}}^2}{{\bm{x}}^{scene}} + {{\bm{b}}^2}} \right) \\ 
\label{eq13}
\end{aligned} 
\end{equation}
The structure of the mixed CCM-CCG is depicted in Fig. \ref{fig7}(c). 

Fusion can be conducted at two levels: feature level and score level, as carried out in \cite{b21}. For score level fusion, an object score in Fig. \ref{fig2} and a scene score in Fig. \ref{fig3} are fed into the trainable fusion module in Fig. \ref{fig1}. In this case, we do not apply softmax on each score vector. For feature level fusion, we use an object feature in Fig. \ref{fig2} and a scene feature in Fig. \ref{fig3} as input features to be fused. This previous study \cite{b21} provides a more detailed explanation.

\section{Experiments}
\label{s4}

The proposed FOSNet is applied to three popular scene recognition datasets, and its performance is compared with that of the previous works. The three scene datasets for the experiment are Places 2 \cite{b20}, SUN 397 \cite{b33}, and MIT indoor 67 \cite{b34}. ImageNet dataset \cite{b19} is also used for the training of ObjectNet.

\subsection{Datasets}
\label{s41}

\textbf{Places 2 dataset} \cite{b20} is the largest dataset for scene recognition. It is an upgraded version of Places 1 \cite{b35}, and it is also the latest of all the scene recognition datasets. This dataset has 365 scene categories; consisting of two versions of datasets: Places365-Challenge dataset and Places365-Standard dataset. Both versions of datasets share the same validation images and only differ in the number of training images. The Places365-Challenge dataset provides 8 million training images, whereas the Places-Standard dataset provides 1.8 million training images.

\textbf{SUN 397 dataset} \cite{b33} was the most popular scene dataset before the Places dataset \cite{b20,b35} was released. This dataset consists of 397 scene categories. Each category has at least 100 different numbers of images. The entire set has a total of 108,754 images. For fair comparison with other methods using this dataset, 10 subsets each of which has 50 training images and 50 validation images per class were used to evaluate the competing methods. The average validation accuracy over the 10 subsets were used as the overall accuracy of each method. 

\textbf{MIT indoor 67 dataset} \cite{b34} is a scene recognition dataset consisting of 67 indoor scene categories, and it comprises a total of 15,620 indoor scene images. All the experiments with the MIT indoor 67 dataset were performed according to the standard evaluation protocol: A subset that has 80 training images and 20 testing images per scene category is used for evaluation. 

\textbf{ImageNet dataset} \cite{b19} is one of the most commonly used datasets for object recognition task, and it consists of 1.2 million object images and 1000 object categories. A number of popular CNN structures were trained in the dataset, which include AlexNet \cite{b18}, ResNet \cite{b23}, DenseNet \cite{b26}, ResNeXt \cite{b27}, SE-Net \cite{b28}, and others \cite{b29,b30,b31}.

\subsection{Implementation Details}
\label{s42}
The FOSNet is comprised of neural networks and it was trained from scratch. All models were trained for 130 epochs. The initial learning rate was 0.15 when the mini-batch size was 256. For different mini-batch sizes, the learning rate was adjusted using the linear scaling rule \cite{b37} to achieve a similar performance. The learning rate was dropped by 0.1 times every 30 epochs. The synchronous stochastic gradient descent with a momentum of 0.9 was used as the optimization method. The training data were augmented by random rescaling, cropped randomly into $224 \times 224$ \cite{b30,b46} and horizontally flipped with a 0.5 chance. The input image was normalized by the per-color mean and standard deviation \cite{b46}. In addition, the data balancing strategy \cite{b11} was adopted for mini-batch sampling \cite{b28}. PlacesNet was trained using Places 2 dataset, and experiments were performed using transfer learning \cite{b38} on other datasets such as SUN 397 and MIT indoor 67.

A hyper-parameter $\gamma$ in Eq. (7) is set to 1. Detailed explanation about $\gamma$ is discussed in Section 5.1. For a backbone network, the SE-ResNeXt-101 model, which is a combination of ResNeXt \cite{b27} with SE-Network \cite{b28}, was used for ObjectNet and PlacesNet in FOSNet. The standard 10-crop testing method \cite{b5} is used for comparison with other methods, and an evaluation measurement is the average classification accuracy of 10 crops.

\subsection{Experimental Results on the Places 2}
\label{s43}
The FOSNet is compared with other scene recognition methods using the validation set of the Places 2 \cite{b20}. The FOSNet is trained using the Places365-Challenge dataset. The comparison with other methods is summarized in Table \ref{tab1}.

\begin{table}[]
\centering
\caption{Comparison with other scene recognition methods on Places 2 \cite{b20} validation set.}
\begin{tabular}{lcc}
\toprule
\multicolumn{1}{c}{\multirow{2}{*}{Methods}}   & \multicolumn{2}{c}{Accuracy (100\%)} \\
\multicolumn{1}{c}{}                           & top-1             & top-5            \\ 
\midrule
Adi-Red \cite{b6}                              & 41.87             & -                \\
Places365-VGG \cite{b20}                       & 55.24             & -                \\
CCM \cite{b21}                                 & 56.82             & 86.92            \\
CNN-SMN \cite{b13}                             & 57.1              & -                \\
SOSF+CFA+GAF \cite{b4}                         & 57.27             & -                \\
Multi-Resolution CNNs \cite{b5}                & 58.3              & 87.3             \\
Places2-365-CNN \cite{b45}                     & 58.93             & 88.52            \\
SE-Resnet-152 \cite{b28}                       & 59.63             & \textbf{88.99}   \\
SE-ResNeXt-101 \cite{b28} - our implementation & 59.10             & 88.44            \\
SE-ResNeXt-101 + SCL                           & 59.80             & 88.66            \\
feature level FOSNet – Sum                     & 59.82             & 88.57            \\
feature level FOSNet – Concatenate             & 60.06             & 88.78            \\
score level FOSNet – CCG                      & 59.92             & 88.69            \\
feature level FOSNet – CCG                    & 60.03             & 88.72            \\
feature level FOSNet – mixed CCM-CCG          & \textbf{60.14}    & 88.86    \\       
\bottomrule
\end{tabular}
\label{tab1}
\end{table}

In Table \ref{tab1}, ``SCL'' denotes a model trained with scene coherence loss and partial convolution \cite{b24} described in Section \ref{s32}. ``Sum'' and ``Concatenate'' denote that the conventional feature fusion methods replace the trainable fusion muddle shown in Fig. \ref{fig1}. ``CCG'' denotes a model using the correlative context gating as revealed in Section \ref{s33}. The entire trainable fusion models use the batch normalization \cite{b32}; The score level or feature level indicates the level at which two kinds of information are combined, as described in Section \ref{s33}.

All the methods listed in Table \ref{tab1} use CNN: Adi-Red \cite{b6}, CCM \cite{b21}, CNN-SMN \cite{b13}, and SOSF + DFA + GAF \cite{b4} used information of the objects which appear in scene images. To obtain the object information, they used the CNN pre-trained on the object recognition dataset. Multi-Resolution CNN \cite{b5} created a super category by considering the label ambiguity of scene categories to train a teacher network. Places365-VGG \cite{b20}, Places2-365-CNN \cite{b45}, and SE-Resnet-152 \cite{b28} used the vanilla CNN architecture and the scene trait was not taken into consideration. In our FOSNet, SE-ResNeXt-101 was used as a backbone network. To demonstrate the competitiveness of SCL, we also conducted experiments using only a SE-ResNeXt-101. When it is trained without considering scene traits, SE-ResNeXt-101 offers lower performance than SE-ResNet-152. However, when SCL is used for training the SE-ResNeXt-101, it outperforms SE-ResNet-152 \cite{b28} with 59.80\%, and this requires lower computation than SE-ResNet-152. When a mixed CCM-CCG model is used, our FOSNet achieves state-of-the-art accuracy of 60.14\% on the Places 2, and it is the first time that the accuracy exceeds 60\% on the dataset.

\subsection{Experimental Results on the SUN 397}
\label{s44}
FOSNet is applied to the SUN 397 dataset \cite{b33}. An average validation accuracy of 10 subsets provided in the dataset is carried out to compare the competing scene recognition methods, and the comparison results are summarized in Table \ref{tab2}.

\begin{table}[]
\centering
\caption{Comparison with other scene recognition methods on SUN 397 \cite{b33}.}
\begin{tabular}{lc}
\toprule
\multicolumn{1}{c}{Methods}           & Accuracy (100\%) \\
\midrule
VS-CNN \cite{b15}                     & 43.14            \\
DAG-CNN \cite{b10}                    & 56.2             \\
Gaze Shifting-CNN+SVM \cite{b12}      & 56.2             \\
MetaObject-CNN \cite{b14}             & 58.11            \\
Places365-VGG-SVM \cite{b20}          & 63.24            \\
Three \cite{b7}                       & 70.17            \\
Hybrid CNN \cite{b8}                  & 70.69            \\
Sparse Representation \cite{b17}      & 71.08            \\
Multi-Resolution CNNs \cite{b5}       & 72.0             \\
CNN-SMN \cite{b13}                    & 72.6             \\
PatchNet \cite{b16}                   & 73.0             \\
SDO \cite{b9}                         & 73.41            \\
Adi-Red \cite{b6}                     & 73.59            \\
SOSF+CFA+GAF \cite{b4}                & \textbf{78.93}   \\
SE-ResNeXt-101 + SCL                  & 76.27            \\
feature level FOSNet – Sum            & 75.32            \\
feature level FOSNet – Concatenate    & 75.57            \\
feature level FOSNet – CCG           & 76.62            \\
feature level FOSNet – mixed CCM-CCG & 77.28           \\
\bottomrule
\end{tabular}
\label{tab2}
\end{table}

The names of competing methods of FOSNet are the same as the names listed in Section 4.3. In this experiment, the object and scene features are combined only at the feature level. Score level fusion is skipped since the PlacesNet is pre-trained with Places 2 dataset but it is applied to SUN 397, making the dimensions of $\bm{x}^{object}$ and $\bm{x}^{scene}$ be different from each other. As shown in Table \ref{tab2}, the model with the mixed CCM-CCG method achieves the highest performance among our methods. Interestingly, when the features are combined using sum or concatenate methods, the performance is degraded from the `SE-ResNeXt-101 + SCL' model which uses only a single scene feature. This shows that a simple increase in the number of features without considering the scene traits can rather hinder scene recognition.

Among the competing methods, FOSNet achieves the second best accuracy of 77.72\%, slightly lower than that of the state-of-the-art SOSF + CFA + GAF method \cite{b4}. Here, it should be noted that it is unfair to directly compare the results of the two methods, considering that SOSF + CFA + GAF \cite{b4} includes YOLOv2 \cite{b41} and 4-directional long short-term memory (LSTM) \cite{b44}. To train YOLOv2, an object detection dataset Object177 \cite{b47} was additionally used. Unlike the dataset used to train ObjectNet, the object detection dataset Object177 includes not only the class labels but also bounding box information for the location of objects in an image, difficult to develop. Furthermore, the method SOSF + CFA + GAF requires much more computation than our method. The FOSNet uses input images of size $224 \times 224$, whereas SOSF + CFA + GAF uses input images of size $608 \times 608$, thereby employing 4-directional LSTM, which is obviously computationally very expensive.

\subsection{Experimental Results on the MIT indoor 67}
\label{s45}
In this subsection, the FOSNet is applied to the validation set of the MIT indoor 67 \cite{b34}, and Table \ref{tab3} presents a comparison result of the scene recognition for MIT 67.

\begin{table}[]
\centering
\caption{Comparison with other scene recognition methods on the MIT 67 \cite{b34} validation set.}
\begin{tabular}{lc}
\toprule
\multicolumn{1}{c}{Methods}           & Accuracy (100\%) \\
\midrule
RBoW \cite{b1}                        & 37.93            \\
DPM+GIST+SP \cite{b2}                 & 43.1             \\
Adi-Red \cite{b6}                     & 73.59            \\
Gaze Shifting-CNN+SVM \cite{b12}      & 75.1             \\
ResNet-152-DFT$^{+}$ \cite{b48}       & 76.5             \\
Places365-VGG-SVM \cite{b20}          & 76.53            \\
DAG-CNN \cite{b10}                    & 77.5             \\
MetaObject-CNN \cite{b14}             & 78.9             \\
VS-CNN \cite{b15}                     & 80.37            \\
Hybrid CNN \cite{b8}                  & 85.97            \\
Three \cite{b7}                       & 86.04            \\
PatchNet \cite{b16}                   & 86.2             \\
CNN-SMN \cite{b13}                    & 86.5             \\
Multi-Resolution CNNs \cite{b5}       & 86.7             \\
SDO \cite{b9}                         & 86.76            \\
Sparse Representation \cite{b17}      & 87.22            \\
SOSF+CFA+GAF \cite{b4}                & 89.51            \\
SE-ResNeXt-101 + SCL                  & 89.10            \\
feature level FOSNet – Sum            & 88.73            \\
feature level FOSNet – Concatenate    & 89.25            \\
feature level FOSNet – CCG           & \textbf{90.37}   \\
feature level FOSNet – mixed CCM-CCG & 90.30           \\
\bottomrule
\end{tabular}
\label{tab3}
\end{table}

In the MIT indoor 67 experiment, only the feature level fusion is performed, and the score level fusion also is skipped based on the same reason as in SUN397. Based on results of Table \ref{tab3}, it seems that the CCG combines the two features more effectively than the existing fusion methods. The result shows the value of the trainable fusion over the heuristic fusion methods.

In Table \ref{tab3}, all the existing methods except RBoW \cite{b1} and DPM+GIST+SP \cite{b2} use CNN. The two methods use the handcraft features. From Table \ref{tab3}, of all the competing methods, the proposed method using CCG offers the best accuracy. In particular, the FOSNet with CCG outperforms SOSF+CFA+GAF \cite{b4} which is the current state-of-the-art method on MIT indoor 67. As a result, our FOSNet with CCG achieves state-of-the-art accuracy of 90.37\% on the MIT indoor 67, and this is the first time that the accuracy exceeds 90\% on the dataset.

\begin{table*}[]
\centering
\caption{Ablation study of SCL using ResNet-18. This experiment is performed on the Places365-Standard dataset \cite{b20}.}
\begin{tabular}{l|cccccc}
\specialrule{1pt}{0pt}{0pt}
\hline
\multicolumn{1}{c|}{Base Model}                       & \multicolumn{6}{c}{ResNet-18}                                                                                               \\ \hline
\multicolumn{1}{c|}{SCL Rate $\left( \gamma \right)$} & \multicolumn{2}{c|}{$0$ (Baseline)}        & \multicolumn{2}{c|}{$10^1$}                & \multicolumn{2}{c}{$10^0$}        \\ \hline
                                                      & top-1 acc & \multicolumn{1}{c|}{top-5 acc} & top-1 acc & \multicolumn{1}{c|}{top-5 acc} & top-1 acc       & top-5 acc       \\ \hline
only SCL                                              & 54.438    & \multicolumn{1}{c|}{84.912}    & 53.775    & \multicolumn{1}{c|}{84.151}    & 54.942          & \textbf{85.074} \\
SCL with Partial Conv                                 & 54.718    & \multicolumn{1}{c|}{84.866}    & 53.688    & \multicolumn{1}{c|}{84.099}    & \textbf{55.090} & 85.027          \\ \hline \hline
\multicolumn{1}{c|}{SCL Rate $\left( \gamma \right)$} & \multicolumn{2}{c|}{$10^{-1}$}             & \multicolumn{2}{c|}{$10^{-2}$}             & \multicolumn{2}{c}{$10^{-3}$}     \\ \hline
                                                      & top-1 acc & \multicolumn{1}{c|}{top-5 acc} & top-1 acc & \multicolumn{1}{c|}{top-5 acc} & top-1 acc       & top-5 acc       \\ \hline
only SCL                                              & 54.770    & \multicolumn{1}{c|}{84.901}    & 54.548    & \multicolumn{1}{c|}{84.715}    & 54.616          & 84.827          \\
SCL with Partial Conv                                 & 54.901    & \multicolumn{1}{c|}{85.038}    & 54.710    & \multicolumn{1}{c|}{84.841}    & 54.877          & 84.929         \\
\specialrule{1pt}{0pt}{0pt}
\end{tabular}
\label{tab4}
\end{table*}

\section{Ablation Study}
\label{s5}
Additional experiments are performed to gain a better understanding of the effects of our proposed SCL and CCG. All ablation studies are performed using the Places365-Standard dataset \cite{b20} for various experiments with fast training. Standard $224 \times 224$ single-crop evaluation is employed, and ResNet-18 and ResNet-50 \cite{b23} are used as the backbone architectures.

\subsection{Analysis on Scene Coherence Loss (SCL)}
\label{s51}
In this subsection, the effects of SCL on scene recognition are demonstrated. In the experiment, only PlacesNet is used and all models in the experiments are trained from scratch for a fair comparison. The results of SCL ablation studies are shown in Tables \ref{tab4} and \ref{tab5}.

\begin{table}[]
\centering
\caption{Ablation study of SCL using ResNet-50. This experiment is performed on the Places365-Standard dataset \cite{b20}.}
\begin{tabular}{l|cccc}
\specialrule{1pt}{0pt}{0pt}
\multicolumn{1}{c|}{Base Model}                       & \multicolumn{4}{c}{ResNet-50}                                           \\ \hline
\multicolumn{1}{c|}{SCL Rate $\left( \gamma \right)$} & \multicolumn{2}{c|}{$0$ (Baseline)}        & \multicolumn{2}{c}{$10^0$} \\ \hline
                                                      & top-1 acc & \multicolumn{1}{c|}{top-5 acc} & top-1 acc    & top-5 acc   \\ \hline
only SCL                                              & 55.888    & \multicolumn{1}{c|}{86.123}    & 56.285       & \textbf{86.288}      \\
SCL with Partial Conv                                 & 56.227    & \multicolumn{1}{c|}{86.099}    & \textbf{56.337}       & 86.195     \\
\specialrule{1pt}{0pt}{0pt}
\end{tabular}
\label{tab5}
\end{table}

\begin{figure}[!t]
\centering
\subfloat[]{
    \includegraphics[width=0.9\linewidth]{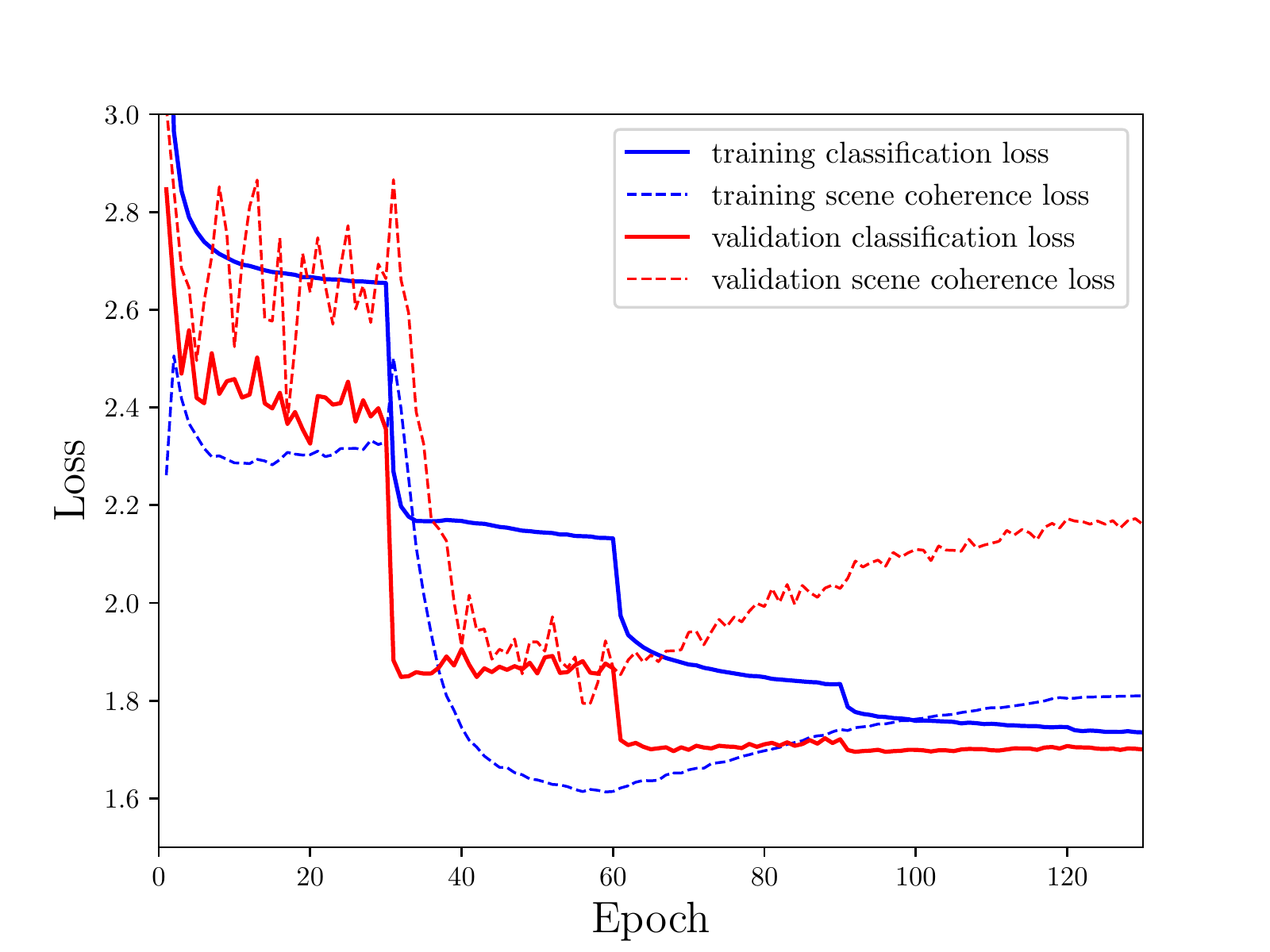}
    \label{fig8a}
}\\
\subfloat[]{
    \includegraphics[width=0.9\linewidth]{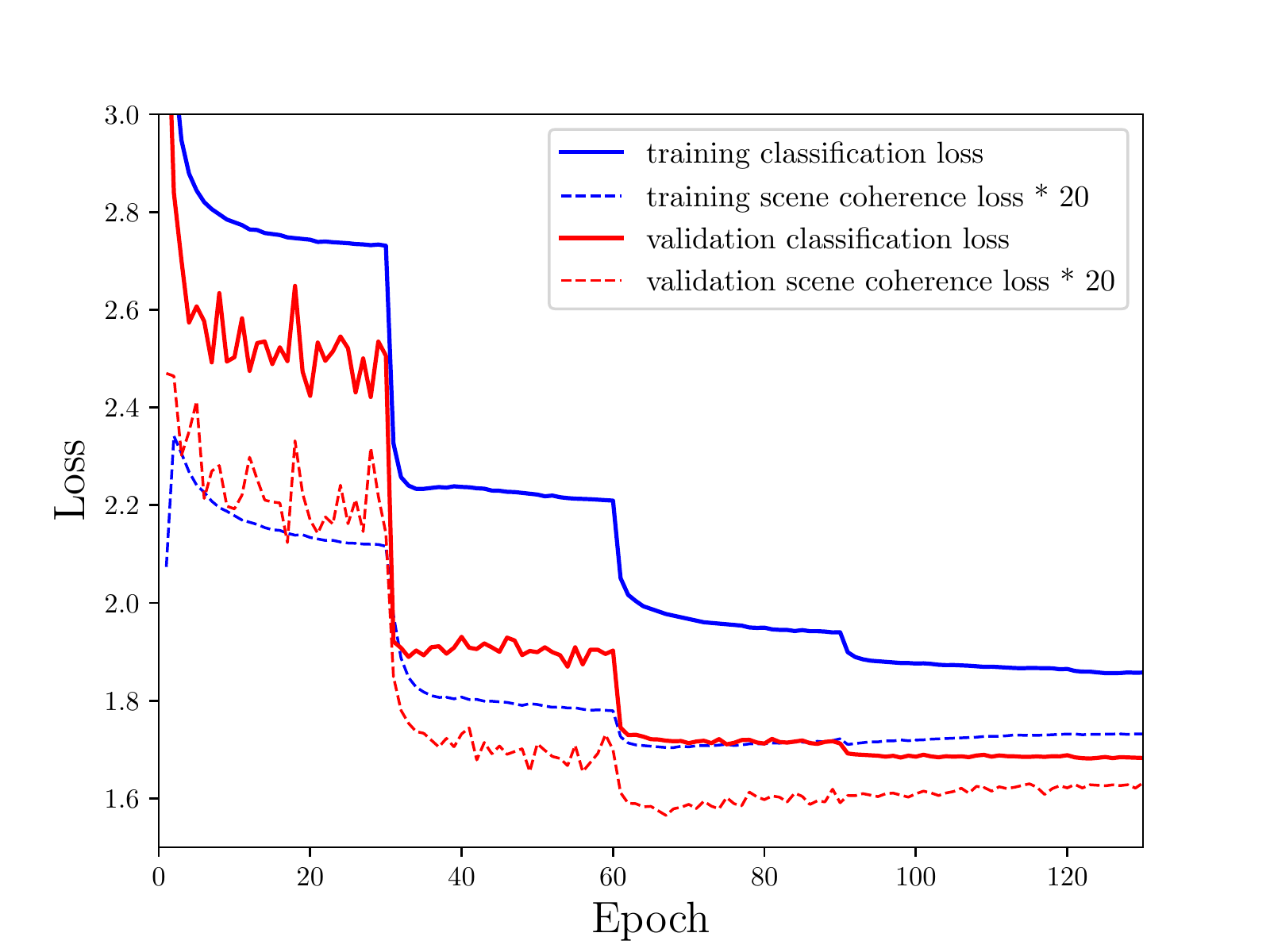}
    \label{fig8b}
}\quad
\caption{Classification loss and SCL curves of ResNet-18 trained (a) with only classification loss and (b) with classification loss and SCL. The blue line denotes the loss of the training set, and the red line denotes the loss of the validation set. The solid line represents classification loss, and the dotted line represents SCL.}
\label{fig8}
\end{figure}

\begin{figure*}[!t]
\centering
\includegraphics[width=0.9\linewidth]{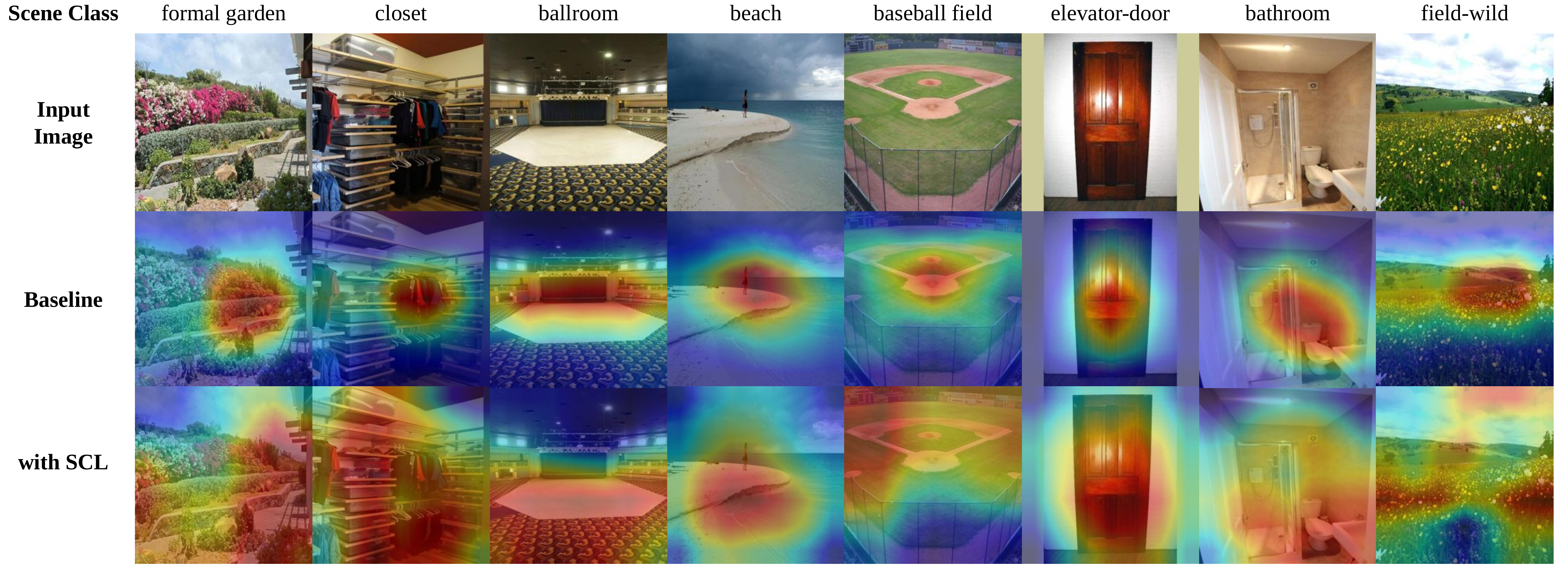}
\caption{The class activation map (CAM) \cite{b25} results using ResNet-18. The ground truth about the scene class of the image is on top of the image. The first row shows the input image. The second row shows the CAM result using ResNet-18 trained without SCL. The third row shows the CAM result using ResNet-18 trained with SCL.}
\label{fig9}
\end{figure*}

\begin{table*}[]
\centering
\caption{Ablation study of CCG on the Places365-Standard dataset \cite{b20}.}
\begin{tabular}{c|l|cc|cc|cc|cc}
\specialrule{1pt}{0pt}{0pt}
\multirow{2}{*}{Fusion Level}  & \multicolumn{1}{c|}{\multirow{2}{*}{Method}} & \multicolumn{2}{c|}{ResNet-18}    & \multicolumn{2}{c|}{ResNet-18 – SCL} & \multicolumn{2}{c|}{ResNet-50}    & \multicolumn{2}{c}{ResNet-50 – SCL} \\
                               & \multicolumn{1}{c|}{}                        & top-1 acc       & top-5 acc       & top-1 acc         & top-5 acc        & top-1 acc       & top-5 acc       & top-1 acc        & top-5 acc        \\ \hline
\multicolumn{1}{l|}{}          & Baseline                                     & 54.438          & 84.912          & 54.942            & 85.074           & 55.888          & 86.123          & 56.285           & 86.288           \\ \hline
\multirow{7}{*}{Feature Level} & Sum                                          & 53.485          & 84.123          & 54.570            & 84.680           & 56.115          & 86.285          & 56.630           & 86.345           \\
                               & Concatenate                                  & 54.701          & 85.071          & 55.164            & 85.145           & 56.230          & \textbf{86.411} & 56.685           & 86.441           \\ \cline{2-10} 
                               & CCM with ReLU \cite{b21}                     & 54.756          & 85.107          & 55.076            & 85.090           & 56.334          & 86.375          & 56.663           & 86.529           \\
                               & CCM-BN with ReLU \cite{b21}                  & 54.786          & 85.181          & 55.129            & 85.230           & 56.269          & 86.395          & 56.726           & \textbf{86.573}  \\ \cline{2-10} 
                               & CCG                                         & 54.575          & 84.888          & 54.907            & 84.921           & 56.060          & 86.186          & 56.469           & 86.233           \\
                               & CCG-BN                                      & \textbf{54.934} & \textbf{85.206} & 55.153            & 85.088           & \textbf{56.367} & 86.395          & 56.729           & 86.397           \\
                               & mixed CCM-CCG-BN                            & 54.504          & 84.997          & 54.959            & 85.132           & 56.060          & 86.373          & \textbf{56.800}  & 86.466           \\ \hline
\multirow{4}{*}{Score Level}   & CCM \cite{b21}                               & 54.477          & 84.869          & 55.055            & 85.107           & 56.096          & 86.233          & 56.581           & 86.315           \\
                               & CCM-BN \cite{b21}                            & 54.600          & 84.979          & 55.104            & 85.126           & 56.110          & 86.238          & 56.690           & 86.343           \\ \cline{2-10} 
                               & CCG                                         & 54.562          & 84.901          & 55.071            & 85.071           & 56.030          & 86.192          & 56.600           & 86.238           \\
                               & CCG-BN                                      & 54.677          & 85.156          & \textbf{55.211}   & \textbf{85.233}  & 56.203          & 86.375          & 56.685           & 86.348          \\
\specialrule{1pt}{0pt}{0pt}
\end{tabular}
\label{tab6}
\end{table*}

In Table \ref{tab4}, accuracy is computed while varying the SCL rate $\left( \gamma \right)$ in Eq. \ref{eq7}. When $\gamma=0$, only classification loss is used as a total loss in Eq. \ref{eq7}. This case is a baseline. When $\gamma \le 1$, accuracy is improved from the baseline in all cases, whereas when $\gamma = 1$, the PlacesNet in FOSNet achieves the best top-1 accuracy. When $\gamma = 10$, the accuracy is degraded, revealing that too much emphasis on SCL is an obstacle to minimizing classification errors. Table \ref{tab5} shows the results of the same experiment using ResNet-50, and similar results are obtained regarding the effects of the SCL with Table \ref{tab4}. The models with SCL always outperform the ones without SCL regardless of which CNN backbone is used. Experimental results regarding the effects of partial convolution \cite{b24} on SCL are given in Tables \ref{tab4} and \ref{tab5}. This partial convolution improves the performance of the baseline, and its effect on the performance is higher when it is combined with SCL. Through the ablation studies, it can be noted that scene recognition performance is improved by using SCL. Since the best performance is obtained when $\gamma = 1$, the value is used to train PlacesNet.

Another experiment is performed to show the validity of the SCL. In Fig. \ref{fig8}(a), the SCL is monitored, and it is unused (not propagated backward) for the training. As shown in Fig. \ref{fig8}(a), the SCL decreases until reaching 60 epochs even when SCL is unused for the training. After 60 epochs in Fig. \ref{fig8}(a), the SCL increases rapidly; the validation loss is almost saturated but the training loss decreases rapidly, revealing that the PlacesNet is overfitted. From the observation, the overfitting in the scene recognition is highly related to SCL. Thus, if the PlacesNet is trained to force the SCL to be reduced, the overfitting of the PlacesNet will be relaxed and its generalization performance will be improved. Fig. \ref{fig8}(b) shows the result when ResNet-18 is trained with SCL. In this case, SCL converges quickly and almost vanishes. Thus, SCL is magnified 20 times for visualization in Fig. \ref{fig8}(b). After 60 epochs in Fig. \ref{fig8}(b), both training and validation errors decrease gradually but consistently, implying that the PlacesNet overfitting is relaxed.

Fig. \ref{fig9} provides the results of class activation map (CAM) \cite{b25} using ResNet-18. The first row shows the input images. Using ResNet-18 trained without and with SCL, the second and third rows show the CAM images, respectively. In the figures, red parts denote the region which is relevant and makes a contribution to the scene classification, whereas the blue parts denote the region that offer no information and no contribution to the scene classification. When trained with SCL, the red region becomes bigger than trained without the SCL. This shows that the region which would be ignored if trained without SCL is fully exploited with SCL. In the entire experiments, the SCL is an effective loss for the scene classification that enables the FOSNet to fully exploit the \textit{sceneness} all over the image.

\subsection{Analysis on Correlative Context Gating (CCG)}
\label{s52}
The proposed feature fusion method CCG is analyzed through experimentation. For a fair comparison, in PlacesNet, CNN models without partial convolution were used for scene feature extraction, and the experimental results are presented in Table \ref{tab6}.

All the models in Table \ref{tab6} are trained from scratch. Compared with the existing fusion methods, such as sum, concatenation or CCM \cite{b21}, our fusion method CCG improves performance in most models regardless of whether SCL is added. Although the fusion by concatenation achieves the best top-5 performance in ResNet-50, the simple fusion delivers limited performances in a new dataset that PlacesNet did not train, as explained in Sections \ref{s44} and \ref{s45}.

\section{Conclusion}
\label{s6}
In this paper, a new scene recognition framework named FOSNet has been proposed, in which the object and the scene information have been combined in a trainable fusion module named CCG. The entire system was trained using SCL, which is a new loss developed for the scene recognition. SCL is based on the unique property of the scene, e.g., the `\textit{sceneness}' spreads and the scene class does not change all over the image. The proposed FOSNet was experimented with three most popular scene recognition datasets, and the state-of-the-art performance is obtained in Places 2 and MIT indoor 67.


%





\ifCLASSOPTIONcaptionsoff
  \newpage
\fi



\bibliographystyle{IEEEtran}
\bibliography{IEEEabrv,My_Collection.bbl}

\begin{thebibliography}{10}
\providecommand{\url}[1]{#1}
\csname url@samestyle\endcsname
\providecommand{\newblock}{\relax}
\providecommand{\bibinfo}[2]{#2}
\providecommand{\BIBentrySTDinterwordspacing}{\spaceskip=0pt\relax}
\providecommand{\BIBentryALTinterwordstretchfactor}{4}
\providecommand{\BIBentryALTinterwordspacing}{\spaceskip=\fontdimen2\font plus
\BIBentryALTinterwordstretchfactor\fontdimen3\font minus
  \fontdimen4\font\relax}
\providecommand{\BIBforeignlanguage}[2]{{%
\expandafter\ifx\csname l@#1\endcsname\relax
\typeout{** WARNING: IEEEtran.bst: No hyphenation pattern has been}%
\typeout{** loaded for the language `#1'. Using the pattern for}%
\typeout{** the default language instead.}%
\else
\language=\csname l@#1\endcsname
\fi
#2}}
\providecommand{\BIBdecl}{\relax}
\BIBdecl

\bibitem{b10}
S.~Yang and D.~Ramanan, ``Multi-scale recognition with {DAG-CNNs},'' in
  \emph{Proceedings of the IEEE International Conference on Computer Vision
  (ICCV)}, 2015, pp. 1215--1223.

\bibitem{b11}
L.~Shen, Z.~Lin, and Q.~Huang, ``Relay backpropagation for effective learning
  of deep convolutional neural networks,'' in \emph{Proceedings of the European
  Conference on Computer Vision (ECCV)}, 2016.

\bibitem{b28}
J.~Hu, L.~Shen, and G.~Sun, ``Squeeze-and-excitation networks,'' in
  \emph{Proceedings of the IEEE Conference on Computer Vision and Pattern
  Recognition (CVPR)}, June 2018.

\bibitem{b48}
J.~Ryu, M.-H. Yang, and J.~Lim, ``Dft-based transformation invariant pooling
  layer for visual classification,'' in \emph{Proceedings of the European
  Conference on Computer Vision (ECCV)}, 2018, pp. 84--99.

\bibitem{b7}
L.~Herranz, S.~Jiang, and X.~Li, ``Scene recognition with cnns: objects, scales
  and dataset bias,'' in \emph{Proceedings of the IEEE Conference on Computer
  Vision and Pattern Recognition (CVPR)}, 2016, pp. 571--579.

\bibitem{b9}
X.~Cheng, J.~Lu, J.~Feng, B.~Yuan, and J.~Zhou, ``Scene recognition with
  objectness,'' \emph{Pattern Recognition}, vol.~74, pp. 474--487, 2018.

\bibitem{b5}
L.~Wang, S.~Guo, W.~Huang, Y.~Xiong, and Y.~Qiao, ``Knowledge guided
  disambiguation for large-scale scene classification with multi-resolution
  cnns,'' \emph{IEEE Transactions on Image Processing}, vol.~26, no.~4, pp.
  2055--2068, 2017.

\bibitem{b3}
A.~Bayat and M.~Pomplun, ``Deriving high-level scene descriptions from deep
  scene cnn features,'' in \emph{Image Processing Theory, Tools and
  Applications (IPTA), 2017 Seventh International Conference on}, 2017.

\bibitem{b21}
H.~Seong, J.~Hyun, H.~Chang, S.~Lee, S.~Woo, and E.~Kim, ``Scene recognition
  via object-to-scene class conversion: end-to-end training,'' in
  \emph{Proceedings of the International Joint Conference on Neural Networks
  (IJCNN)}, July 2019.

\bibitem{b19}
O.~Russakovsky, J.~Deng, H.~Su, J.~Krause, S.~Satheesh, S.~Ma, Z.~Huang,
  A.~Karpathy, A.~Khosla, M.~Bernstein \emph{et~al.}, ``Imagenet large scale
  visual recognition challenge,'' \emph{International Journal of Computer
  Vision}, vol. 115, no.~3, pp. 211--252, 2015.

\bibitem{b42}
M.~Everingham, L.~Van~Gool, C.~K. Williams, J.~Winn, and A.~Zisserman, ``The
  pascal visual object classes (voc) challenge,'' \emph{International Journal
  of Computer Vision}, vol.~88, no.~2, pp. 303--338, 2010.

\bibitem{b43}
T.-Y. Lin, M.~Maire, S.~Belongie, J.~Hays, P.~Perona, D.~Ramanan,
  P.~Doll{\'a}r, and C.~L. Zitnick, ``Microsoft coco: Common objects in
  context,'' in \emph{Proceedings of the European Conference on Computer Vision
  (ECCV)}.\hskip 1em plus 0.5em minus 0.4em\relax Springer, 2014, pp. 740--755.

\bibitem{b6}
Z.~Zhao and M.~Larson, ``From volcano to toyshop: Adaptive discriminative
  region discovery for scene recognition,'' in \emph{Proceedings of the 26th
  ACM International Conference on Multimedia (ACM MM)}, 2018, pp. 1760--1768.

\bibitem{b13}
X.~Song, S.~Jiang, and L.~Herranz, ``Multi-scale multi-feature context modeling
  for scene recognition in the semantic manifold,'' \emph{IEEE Transactions on
  Image Processing}, vol.~26, no.~6, pp. 2721--2735, 2017.

\bibitem{b14}
R.~Wu, B.~Wang, W.~Wang, and Y.~Yu, ``Harvesting discriminative meta objects
  with deep cnn features for scene classification,'' in \emph{Proceedings of
  the IEEE International Conference on Computer Vision (ICCV)}, 2015, pp.
  1287--1295.

\bibitem{b15}
J.~Shi, H.~Zhu, S.~Yu, W.~Wu, and H.~Shi, ``Scene categorization model using
  deep visually sensitive features,'' \emph{IEEE Access}, 2019.

\bibitem{b2}
M.~Pandey and S.~Lazebnik, ``Scene recognition and weakly supervised object
  localization with deformable part-based models,'' in \emph{Proceedings of the
  IEEE International Conference on Computer Vision (ICCV)}, 2011.

\bibitem{b1}
S.~N. Parizi, J.~G. Oberlin, and P.~F. Felzenszwalb, ``Reconfigurable models
  for scene recognition,'' in \emph{Proceedings of the IEEE Conference on
  Computer Vision and Pattern Recognition (CVPR)}, 2012, pp. 2775--2782.

\bibitem{b12}
X.~Sun, L.~Zhang, Z.~Wang, J.~Chang, Y.~Yao, P.~Li, and R.~Zimmermann, ``Scene
  categorization using deeply learned gaze shifting kernel,'' \emph{IEEE
  Transactions on Cybernetics}, 2019.

\bibitem{b4}
N.~Sun, W.~Li, J.~Liu, G.~Han, and C.~Wu, ``Fusing object semantics and deep
  appearance features for scene recognition,'' \emph{IEEE Transactions on
  Circuits and Systems for Video Technology}, 2018.

\bibitem{b8}
G.-S. Xie, X.-Y. Zhang, S.~Yan, and C.-L. Liu, ``Hybrid cnn and
  dictionary-based models for scene recognition and domain adaptation,''
  \emph{IEEE Transactions on Circuits and Systems for Video Technology},
  vol.~27, no.~6, pp. 1263--1274, 2017.

\bibitem{b16}
Z.~Wang, L.~Wang, Y.~Wang, B.~Zhang, and Y.~Qiao, ``Weakly supervised
  patchnets: Describing and aggregating local patches for scene recognition,''
  \emph{IEEE Transactions on Image Processing}, vol.~26, no.~4, pp. 2028--2041,
  2017.

\bibitem{b17}
G.~Nascimento, C.~Laranjeira, V.~Braz, A.~Lacerda, and E.~R. Nascimento, ``A
  robust indoor scene recognition method based on sparse representation,'' in
  \emph{Iberoamerican Congress on Pattern Recognition}.\hskip 1em plus 0.5em
  minus 0.4em\relax Springer, 2017, pp. 408--415.

\bibitem{b49}
C.~Cortes and V.~Vapnik, ``Support-vector networks,'' \emph{Machine learning},
  vol.~20, no.~3, pp. 273--297, 1995.

\bibitem{b23}
K.~He, X.~Zhang, S.~Ren, and J.~Sun, ``Deep residual learning for image
  recognition,'' in \emph{Proceedings of the IEEE Conference on Computer Vision
  and Pattern Recognition (CVPR)}, 2016, pp. 770--778.

\bibitem{b26}
G.~Huang, Z.~Liu, L.~Van Der~Maaten, and K.~Q. Weinberger, ``Densely connected
  convolutional networks,'' in \emph{Proceedings of the IEEE Conference on
  Computer Vision and Pattern Recognition (CVPR)}, July 2017.

\bibitem{b27}
S.~Xie, R.~Girshick, P.~Doll{\'a}r, Z.~Tu, and K.~He, ``Aggregated residual
  transformations for deep neural networks,'' in \emph{Proceedings of the IEEE
  Conference on Computer Vision and Pattern Recognition (CVPR)}, 2017, pp.
  1492--1500.

\bibitem{b20}
B.~Zhou, A.~Lapedriza, A.~Khosla, A.~Oliva, and A.~Torralba, ``Places: A 10
  million image database for scene recognition,'' \emph{IEEE Transactions on
  Pattern Analysis and Machine Intelligence}, vol.~40, no.~6, pp. 1452--1464,
  2018.

\bibitem{b25}
B.~Zhou, A.~Khosla, A.~Lapedriza, A.~Oliva, and A.~Torralba, ``Learning deep
  features for discriminative localization,'' in \emph{Proceedings of the IEEE
  Conference on Computer Vision and Pattern Recognition (CVPR)}, 2016, pp.
  2921--2929.

\bibitem{b24}
G.~Liu, K.~J. Shih, T.~Wang, F.~A. Reda, K.~Sapra, Z.~Yu, A.~Tao, and
  B.~Catanzaro, ``Partial convolution based padding,'' \emph{arXiv preprint
  arXiv:1811.11718}, 2018.

\bibitem{b22}
A.~Miech, I.~Laptev, and J.~Sivic, ``Learnable pooling with context gating for
  video classification,'' \emph{arXiv preprint arXiv:1706.06905}, 2017.

\bibitem{b32}
S.~Ioffe and C.~Szegedy, ``Batch normalization: Accelerating deep network
  training by reducing internal covariate shift,'' in \emph{Proceedings of the
  International Conference on Machine Learning (ICML)}, 2015.

\bibitem{b33}
J.~Xiao, J.~Hays, K.~A. Ehinger, A.~Oliva, and A.~Torralba, ``Sun database:
  Large-scale scene recognition from abbey to zoo,'' in \emph{Proceedings of
  the IEEE Conference on Computer Vision and Pattern Recognition (CVPR)}, 2010,
  pp. 3485--3492.

\bibitem{b34}
A.~Quattoni and A.~Torralba, ``Recognizing indoor scenes,'' in
  \emph{Proceedings of the IEEE Conference on Computer Vision and Pattern
  Recognition (CVPR)}, 2009, pp. 413--420.

\bibitem{b35}
B.~Zhou, A.~Lapedriza, J.~Xiao, A.~Torralba, and A.~Oliva, ``Learning deep
  features for scene recognition using places database,'' in \emph{Proceedings
  of the Advances in Neural Information Processing Systems (NIPS)}, 2014, pp.
  487--495.

\bibitem{b18}
A.~Krizhevsky, I.~Sutskever, and G.~E. Hinton, ``Imagenet classification with
  deep convolutional neural networks,'' in \emph{Proceedings of the Advances in
  Neural Information Processing Systems (NIPS)}, 2012, pp. 1097--1105.

\bibitem{b29}
K.~Simonyan and A.~Zisserman, ``Very deep convolutional networks for
  large-scale image recognition,'' in \emph{Proceedings of the International
  Conference on Learning Representations (ICLR)}, 2015.

\bibitem{b30}
C.~Szegedy, W.~Liu, Y.~Jia, P.~Sermanet, S.~Reed, D.~Anguelov, D.~Erhan,
  V.~Vanhoucke, and A.~Rabinovich, ``Going deeper with convolutions,'' in
  \emph{Proceedings of the IEEE Conference on Computer Vision and Pattern
  Recognition (CVPR)}, 2015.

\bibitem{b31}
C.~Szegedy, V.~Vanhoucke, S.~Ioffe, J.~Shlens, and Z.~Wojna, ``Rethinking the
  inception architecture for computer vision,'' in \emph{Proceedings of the
  IEEE Conference on Computer Vision and Pattern Recognition (CVPR)}, 2016, pp.
  2818--2826.

\bibitem{b37}
P.~Goyal, P.~Doll{\'a}r, R.~Girshick, P.~Noordhuis, L.~Wesolowski, A.~Kyrola,
  A.~Tulloch, Y.~Jia, and K.~He, ``Accurate, large minibatch sgd: training
  imagenet in 1 hour,'' \emph{arXiv preprint arXiv:1706.02677}, 2017.

\bibitem{b46}
S.~Gross and M.~Wilber, ``Training and investigating residual nets,''
  \url{https://github.com/facebook/fb.resnet.torch}, 2016.

\bibitem{b38}
S.~J. Pan, Q.~Yang \emph{et~al.}, ``A survey on transfer learning,'' \emph{IEEE
  Transactions on knowledge and data engineering}, vol.~22, no.~10, pp.
  1345--1359, 2010.

\bibitem{b45}
L.~Shen, Z.~Lin, G.~Sun, and J.~Hu, ``Places401 and places365 models,''
  \url{https://github.com/lishen-shirley/Places2-CNNs}, 2016.

\bibitem{b41}
J.~Redmon and A.~Farhadi, ``Yolo9000: better, faster, stronger,'' in
  \emph{Proceedings of the IEEE Conference on Computer Vision and Pattern
  Recognition (CVPR)}, 2017, pp. 7263--7271.

\bibitem{b44}
S.~Hochreiter and J.~Schmidhuber, ``Long short-term memory,'' \emph{Neural
  computation}, vol.~9, no.~8, pp. 1735--1780, 1997.

\bibitem{b47}
L.-J. Li, H.~Su, L.~Fei-Fei, and E.~P. Xing, ``Object bank: A high-level image
  representation for scene classification \& semantic feature sparsification,''
  in \emph{Proceedings of the Advances in Neural Information Processing Systems
  (NIPS)}, 2010, pp. 1378--1386.

\end{thebibliography}

\begin{IEEEbiography}[{\includegraphics[width=1in,height=1.25in,clip,keepaspectratio]{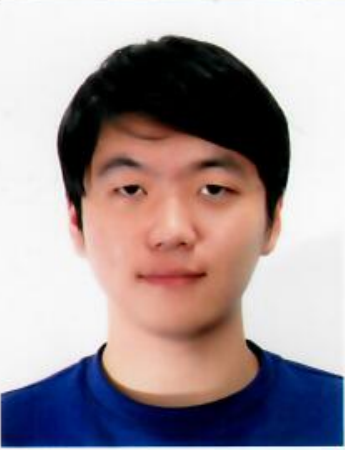}}]{Hongje Seong} received the BS degree in electrical and electronic engineering from Yonsei University, Seoul, Korea, in 2018. He is a graduate student of the combined masters and doctoral degree programs at Yonsei University. He has studied computer vision, machine learning and deep learning.
\end{IEEEbiography}

\begin{IEEEbiography}[{\includegraphics[width=1in,height=1.25in,clip,keepaspectratio]{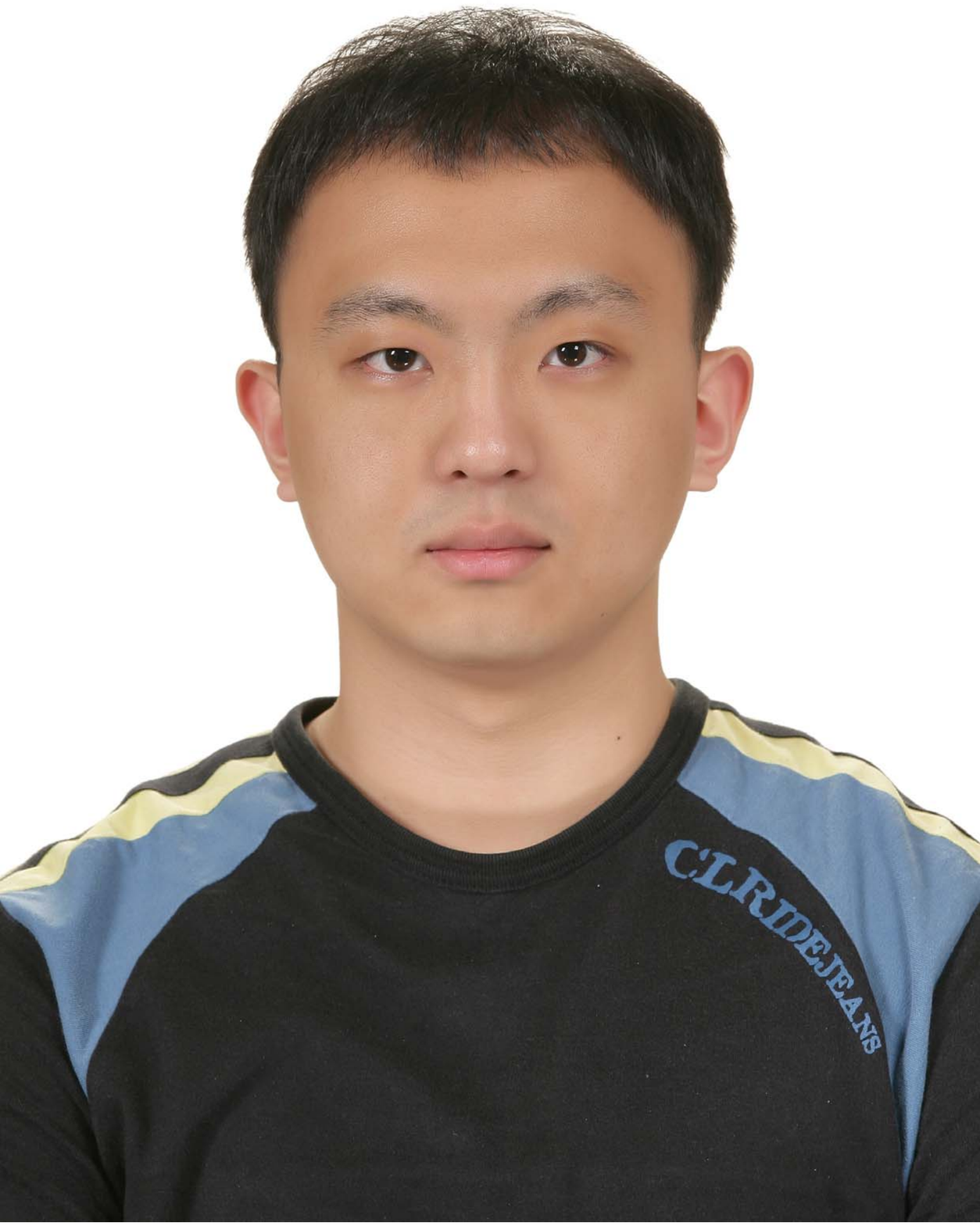}}]{Junhyuk Hyun} received the BS degree in electrical and electronic engineering from Yonsei University, Seoul, Korea, in 2014. He is a graduate student of the combined masters and doctoral degree programs at Yonsei University. He has studied computer vision, machine learning and deep learning.
\end{IEEEbiography}

\begin{IEEEbiography}[{\includegraphics[width=1in,height=1.25in,clip,keepaspectratio]{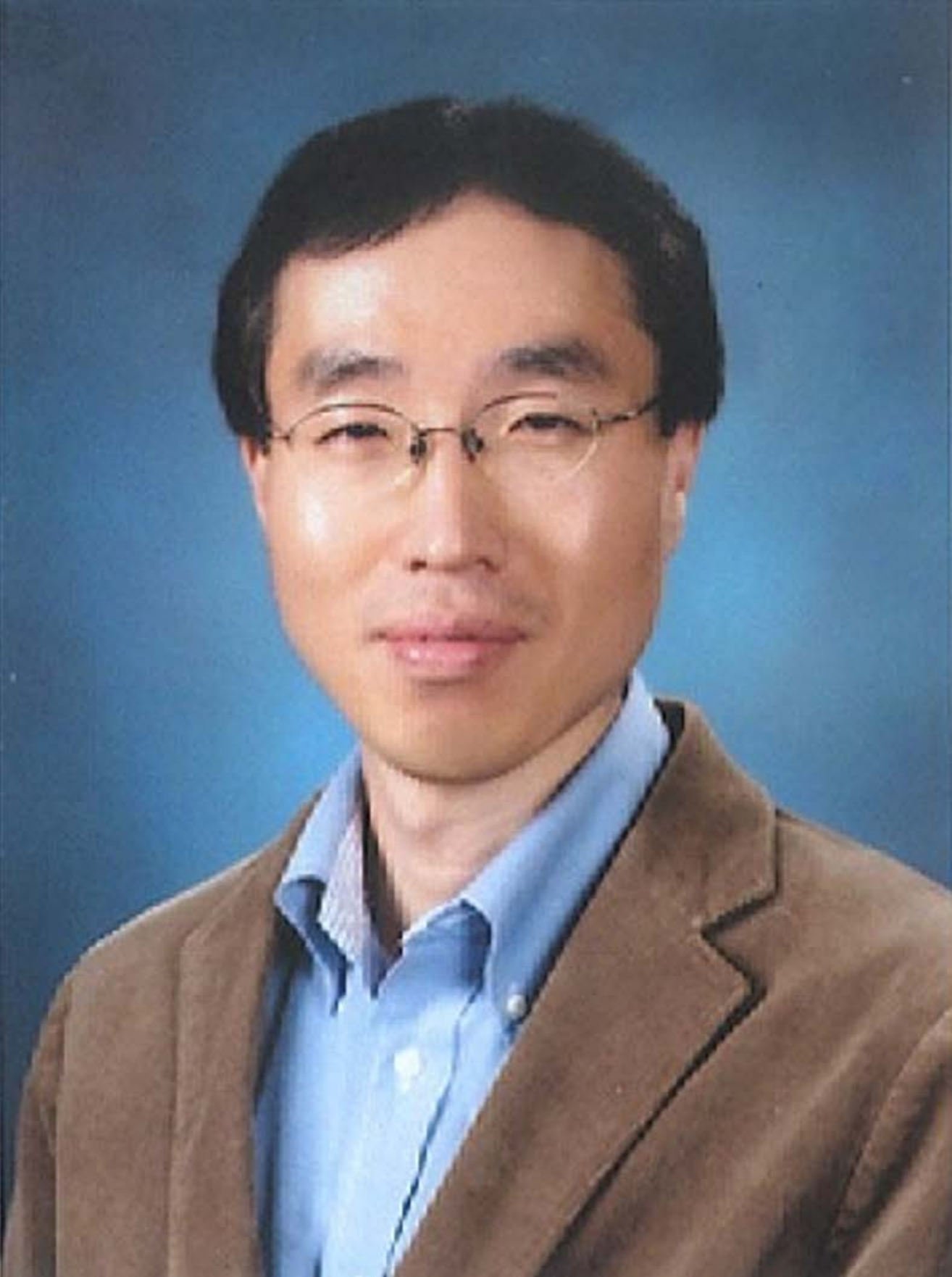}}]{Euntai Kim} was born in Seoul, Korea in 1970. He received B.S., M.S., and Ph.D. degrees in Electronic Engineering, all from Yonsei University, Seoul, Korea, in 1992, 1994, and 1999, respectively. From 1999 to 2002, he was a Full-Time Lecturer in the Department of Control and Instrumentation Engineering, Hankyong National University, Kyonggi-do, Korea.

Since 2002, he has been with the faculty of the School of Electrical and Electronic Engineering, Yonsei University, where he is currently a Professor. He was also a visiting researcher at the Berkeley Initiative in Soft Computing, University of California, Berkeley, CA, USA, in 2008. His current research interests include computational intelligence, statistical machine learning and deep learning and their application to intelligent robotics, autonomous vehicles, and robot vision.
\end{IEEEbiography}




\end{document}